\documentclass[sigconf,natbib=true,screen]{acmart}
\usepackage{xspace,balance,tabularx,multirow}
\usepackage{flushend}
\usepackage{tikz}
\usepackage{pgfplots}
\usetikzlibrary{patterns}
\usepackage{subfig}
\usepackage[ruled, vlined, linesnumbered]{algorithm2e}
\usepackage{xcolor}
\usepackage{colortbl}
\usepackage{bbold}
\SetKwComment{Comment}{$\triangleright$\ }{}
\usepackage{enumitem}
\usepackage{tablefootnote}
\usepackage{upgreek,textgreek}
\usepackage{pifont}%
\usepackage[noabbrev]{cleveref}
\usepackage{titlecaps}
\usepackage{lipsum}

\usepackage{hyperref}

\usepackage{graphicx}
\usepackage{amsmath}
\usepackage{booktabs}
\usepackage{caption}
\usepackage{adjustbox}

\pgfplotsset{every tick label/.append style={font=\tiny}}

\newlength{\starsize}
\newlength{\starspread}
\tikzset{starsize/.code={\setlength{\starsize}{#1}},
         starspread/.code={\setlength{\starspread}{#1}}}
\tikzset{starsize=1mm,
         starspread=3mm}
\pgfdeclarepatternformonly[\starspread,\starsize]%
  {my fivepointed stars}%
  {\pgfpointorigin}%
  {\pgfqpoint{\starspread}{\starspread}}%
  {\pgfqpoint{\starspread}{\starspread}}%
  {%
   \pgftransformshift{\pgfqpoint{\starsize}{\starsize}}
   \pgfpathmoveto{\pgfqpointpolar{18}{\starsize}}
   \pgfpathlineto{\pgfqpointpolar{162}{\starsize}}
   \pgfpathlineto{\pgfqpointpolar{306}{\starsize}}
   \pgfpathlineto{\pgfqpointpolar{90}{\starsize}}
   \pgfpathlineto{\pgfqpointpolar{234}{\starsize}}
   \pgfpathclose%
   \pgfusepath{fill}
  }

\makeatletter
\newcommand*\bigcdot{\mathpalette\bigcdot@{.5}}
\newcommand*\bigcdot@[2]{\mathbin{\vcenter{\hbox{\scalebox{#2}{$\m@th#1\bullet$}}}}}
\makeatother

\newcommand{\stitle}[1]{\vspace*{0.5em}\noindent{\bf #1.\/}}

\def\header{\vspace{1mm} \noindent}

\newcommand{\ccate}{\mathbin\Vert\xspace}

\newcommand{\vcate}{\mathbin\nparallel\xspace}

\newcommand{\U}{\mathcal{U}\xspace}
\newcommand{\V}{\mathcal{I}\xspace}
\newcommand{\G}{\mathcal{G}\xspace}
\newcommand{\N}{\mathcal{N}\xspace}

\newcommand{\EDG}{\mathcal{E}\xspace}
\newcommand{\DOC}{\mathcal{T}\xspace}
\newcommand{\NAM}{\widehat{\boldsymbol{A}}\xspace}
\newcommand{\dvec}{\boldsymbol{d}\xspace}
\newcommand{\Ss}{\mathcal{S}\xspace}

\newcommand{\WM}{\boldsymbol{W}\xspace}
\newcommand{\AM}{\boldsymbol{A}\xspace}

\newcommand{\DM}{\textsf{diag}(\dvec)\xspace}
\newcommand{\DUM}{{\textsf{diag}\left(\dvec^{(\textrm{u})}\right)}\xspace}
\newcommand{\DVM}{{\textsf{diag}\left(\dvec^{(\textrm{i})}\right)}\xspace}
\newcommand{\EUM}{{\boldsymbol{E}^{(\textrm{u})}}\xspace}
\newcommand{\EVM}{{\boldsymbol{E}^{(\textrm{i})}}\xspace}

\newcommand{\PUM}{{\boldsymbol{P}^{(\textrm{u})}}\xspace}
\newcommand{\PVM}{{\boldsymbol{P}^{(\textrm{i})}}\xspace}

\newcommand{\IM}{\boldsymbol{I}\xspace}
\newcommand{\SM}{\boldsymbol{S}\xspace}

\newcommand{\PM}{\boldsymbol{P}\xspace}

\newcommand{\YM}{\boldsymbol{Y}\xspace}
\newcommand{\XM}{\boldsymbol{X}\xspace}
\newcommand{\LM}{\boldsymbol{L}\xspace}
\newcommand{\UM}{\boldsymbol{U}\xspace}
\newcommand{\VM}{\boldsymbol{V}\xspace}
\newcommand{\HM}{\boldsymbol{H}\xspace}
\newcommand{\KM}{\boldsymbol{K}\xspace}

\newcommand{\BM}{\boldsymbol{B}\xspace}

\newcommand{\TM}{\boldsymbol{T}\xspace}
\newcommand{\EM}{\boldsymbol{E}\xspace}
\newcommand{\GaM}{\boldsymbol{\Gamma}\xspace}
\newcommand{\DeM}{\boldsymbol{\Delta}\xspace}
\newcommand{\LaM}{\boldsymbol{\Lambda}\xspace}
\newcommand{\PhiM}{\boldsymbol{\Phi}\xspace}
\newcommand{\PsiM}{\boldsymbol{\Psi}\xspace}

\newcommand{\ZM}{\boldsymbol{Z}\xspace}

\newcommand{\QM}{\boldsymbol{Q}\xspace}

\newcommand{\algo}{\textsf{SAFT}\xspace}
\newcommand{\algognn}{\textsf{SAFT} (LGA)\xspace}
\newcommand{\algogau}{\textsf{SAFT} (GAU)\xspace}

\newcommand{\eat}[1]{}

\newenvironment{customlegend}[1][]{%
    \begingroup
    \csname pgfplots@init@cleared@structures\endcsname
    \pgfplotsset{#1}%
}{%
    \csname pgfplots@createlegend\endcsname
    \endgroup
}%

\def\addlegendimage{\csname pgfplots@addlegendimage\endcsname}

\makeatletter
\newcommand\footnoteref[1]{\protected@xdef\@thefnmark{\ref{#1}}\@footnotemark}
\makeatother

\let\oldnl\nl%
\newcommand{\nonl}{\renewcommand{\nl}{\let\nl\oldnl}}%

\SetKwComment{Comment}{/* }{ */}

\makeatletter 
\g@addto@macro{\@algocf@init}{\SetKwInOut{Parameter}{Parameters}} 
\makeatother

\definecolor{myred}{HTML}{fd7f6f}
\definecolor{myred_new}{HTML}{D8D8D8}
\definecolor{myred_new2}{HTML}{D7191C}
\definecolor{myblue}{HTML}{7eb0d5}
\definecolor{mygreen}{HTML}{b2e061}
\definecolor{mypurple}{HTML}{bd7ebe}
\definecolor{myorange}{HTML}{ffb55a}
\definecolor{myyellow}{HTML}{ffee65}
\definecolor{mypurple2}{HTML}{beb9db}
\definecolor{mypink}{HTML}{fdcce5}
\definecolor{mycyan}{HTML}{8bd3c7}

\definecolor{myblue2}{HTML}{115f9a}
\definecolor{myred2}{HTML}{c23728}

\AtBeginDocument{%
  \providecommand\BibTeX{{%
    \normalfont B\kern-0.5em{\scshape i\kern-0.25em b}\kern-0.8em\TeX}}}

\setcopyright{acmcopyright}
\copyrightyear{2018}
\acmYear{2018}
\acmDOI{XXXXXXX.XXXXXXX}

\acmPrice{15.00}
\acmISBN{978-1-4503-XXXX-X/18/06}

\acmSubmissionID{1389}

\begin{document}

\title{\algo: Structure-aware Transformers for Textual Interaction Classification}
\subtitle{Technical Report}
\author{Hongtao Wang}
\affiliation{%
  \institution{Hong Kong Baptist University}
  \country{Hong Kong SAR, China}
}
\email{cshtwang@comp.hkbu.edu.hk}
\author{Renchi Yang}
\affiliation{%
  \institution{Hong Kong Baptist University}
  \country{Hong Kong SAR, China}
}
\email{renchi@hkbu.edu.hk}
\author{Hewen Wang}
\affiliation{%
  \institution{National University of Singapore}
  \country{Singapore, Singapore}
}
\email{wanghewen@u.nus.edu}
\author{Haoran Zheng}
\affiliation{%
  \institution{Hong Kong Baptist University}
  \country{Hong Kong SAR, China}
}
\email{cshrzheng@comp.hkbu.edu.hk}
\author{Jianliang Xu}
\affiliation{%
  \institution{Hong Kong Baptist University}
  \country{Hong Kong SAR, China}
}
\email{xujl@comp.hkbu.edu.hk}

\settopmatter{printfolios=true}

\renewcommand{\shortauthors}{Wang et al.}

\begin{abstract}
Textual interaction networks (TINs) are an omnipresent data structure used to model the interplay between users and items on e-commerce websites, social networks, etc., where each interaction is associated with a text description. Classifying such textual interactions (TIC) finds extensive use in detecting spam reviews in e-commerce, fraudulent transactions in finance, and so on.
Existing TIC solutions either (i) fail to capture the rich text semantics due to the use of context-free text embeddings, and/or (ii) disregard the bipartite structure and node heterogeneity of TINs, leading to compromised TIC performance.
In this work, we propose \algo, a new architecture that integrates language- and graph-based modules for the effective fusion of textual and structural semantics in the representation learning of interactions.
In particular, 
{\em line graph attention} (LGA)/{\em gated attention units} (GAUs) and {\em pretrained language models} (PLMs) are capitalized on to model the interaction-level and token-level signals, which are further coupled via the proxy token in an iterative and contextualized fashion. 
Additionally, an efficient and theoretically-grounded approach is developed to encode the local and global topology information pertaining to interactions into structural embeddings. The resulting embeddings not only inject the structural features underlying TINs into the textual interaction encoding but also facilitate the design of graph sampling strategies.
Extensive empirical evaluations on multiple real TIN datasets demonstrate the superiority of \algo over the state-of-the-art baselines in TIC accuracy. 
\end{abstract}



\maketitle

\section{Introduction}
{\em Textual interaction networks} (TINs) are a data model characterizing the interactive behaviors between two sets of heterogeneous entities (i.e., users and items), wherein each interaction is accompanied by a textual description.
Such data structures are prevalent in various real-world scenarios, e.g., users' reviews of businesses/products, course feedback from students, transactions between consumers and merchants on e-commerce websites, and user posts on topics in social media, etc.
Given a TIN $\G$ and partially observed labels for interactions, the goal of {\em textual interaction classification} (hereinafter TIC) is to predict the categories of the remaining textual interactions.
In practice, TIC has seen a wide range of applications in spam review detection~\cite{yu2023mrfs}, 
identifying fraudulent financial transactions~\cite{wang2022review},
sentiment analysis or stance detection in social networks~\cite{vishal2016sentiment,aldayel2021stance}, information retrieval~\cite{mao2020item}, and many others \cite{fathony2023interaction,pandey2019comprehensive,wang2021bipartite}.

One simple treatment for TIC is to apply natural language processing techniques to the textual data of the interactions, regardless of the graph topology of TINs, leading to sub-optimal results.
Another line of research formulates TIC as an edge classification problem~\cite{aggarwal2016edge} in the graph, which seeks to exploit the graph structure for label prediction. 
Over the past few years, along this line, considerable efforts have been invested towards learning predictive representations (a.k.a. embeddings) for edges from their attributes and structural semantics underlying the interaction networks. 
The majority of them~\cite{jo2021edge,jiang2019censnet,kim2019edge} harness the powerful ability of {\em graph neural networks} (GNNs)~\cite{wu2020comprehensive} in fusing node/edge attributes and graph structure for effective representation learning, while others~\cite{bielak2022attre2vec,wang2020edge2vec,wang2023efficient} resort to random walks or deep auto-encoders to extract topological features.
These works fail on TINs as they are primarily designed for unipartite networks, which overlook the bipartite nature and node heterogeneity of TINs.
As a remedy, \citet{wang2024effective} extend the {\em message passing} scheme for homogeneous nodes in GNNs to edges by employing two heterogeneous sets of nodes in TINs as intermediaries, respectively.

However, the foregoing methodology suffers from limited model capability
due to a fundamental deficiency: the use of bag-of-words and context-free text embeddings.
To be more specific, these models rely on a cascaded architecture in which a front-mounted stage is adopted to transform textual descriptions into shallow text embeddings as edge attributes using TF-IDF~\cite{robertson1994some} or Word2Vec~\cite{mikolov2013distributed} before running edge-aware GNNs. Such a pipeline is efficient but inherently limited in fully capturing contextualized text semantics and extracting task-relevant features.

Inspired by the remarkable success of {\em pretrained language models} (PLMs) (e.g., BERT~\cite{devlin2018bert}) in encoding contextual and linguistic semantics in text, recent advances~\cite{zhao2022learning,yang2021graphformers} in textual graph representation learning focus on integrating PLMs into GNNs for co-training, thereby mutually enhancing text and node embeddings.
Despite of their improvements, they cannot be readily applied for TIC as they are specially catered for node-wise tasks.
In a recent work~\cite{jin2022edgeformers}, Jin et al. employ PLMs to model edge text for textual edge classification by simply injecting randomly generated node tokens into each Transformer~\cite{vaswani2017attention} layer inside the PLMs. Therein, the graph structures, especially the unique bipartite characteristics of the TIN $\G$, are largely neglected, which, in turn, compromises the embedding utility. 
In sum, existing solutions to TIC fall short of the exploitation of either textual semantics or structural information in TINs for TIC, and it remains unclear how to fuse both in a contextualized and unified manner.

\stitle{Present Work}
To bridge this gap, this paper presents \algo, a \underline{S}tructure-\underline{A}ware Trans\underline{F}ormer for \underline{T}extual interaction classification in TINs. 
The inspiration of \algo stems from the message passing interpretation of the attention mechanism~\cite{vaswani2017attention} in Transformers, in which token-to-token graphs (i.e., attention matrices) are learned for feature propagation between text tokens within each textual interaction.
Ideally, the message passing should go beyond text tokens to the interaction level (i.e., edge-level), so as to enable (i) the feature aggregation from users and items via topological connections, and (ii) deep interplay between the signals from macroscopic (interactions) and microscopic (text tokens) views.

More concretely, at the macroscopic level, \algo aims to obtain attention coefficients for users and items that aid the interaction-user and -item message passing, respectively. 
In lieu of opting for fully learned attention weights for all user/item/interaction pairs, \algo incorporates the bipartite topology of TINs into our LGA ({\em line graph attention}) and GAUs ({\em gated attention unit})~\cite{hua2022transformer} for a linear-complexity but efficacious feature aggregation.
To facilitate the deep coupling of the macroscopic and microscopic modules, \algo includes a proxy token in each input sequence in Transformers, which underpins the roles of aggregating semantics from text tokens, integrating the textual, user-wise, and item-wise features of each textual interaction, and cross-layer information transmission.

On top of that, to adequately involve the structural signals and patterns of TINs in the process of textual interaction encoding, in each input sequence in Transformers, we introduce two structural identity tokens using our structural embeddings that encode the global centrality ({\em spanning centrality}~\cite{mavroforakis2015spanning}) and local connectivity (edge-to-edge {\em resistance distance}~\cite{klein1993resistance}) of interactions in $\G$.
Particularly, a simple but theoretically-grounded approach is devised to construct the centrality and distance embeddings through a fast truncated {\em singular value decomposition} (SVD) of the input TIN topology.
Furthermore, we develop two graph sampling strategies to cherry-pick a handful of adjacent interactions/edges for efficient interaction-user/-item message passing and model training based on the distance and centrality embeddings.
Our extensive experiments comparing \algo against 17 baselines on 8 real TIN datasets reveal that \algo consistently achieves superior prediction accuracy in terms of TIC tasks.

\section{Related Work}
\subsection{Semi-Supervised Edge Classification}

In a pioneering work~\cite{aggarwal2016edge}, the unknown labels of edges are inferred from known ones based on their structural similarities. Subsequent research has predominantly focused on generating embeddings for edges. One line of work employs shallow embedding techniques. AttrE2vec~\cite{bielak2022attre2vec} uses random walk to achieve message passing, while Edge2vec~\cite{wang2020edge2vec} combines the deep autoencoder and skip-gram model. TER+AER~\cite{wang2023efficient} generates
high-quality edge representation vectors based on the graph structure surrounding edges and edge attributes. Recent methods like TopoEdge~\cite{cheng2024edge} which further enhances edge embeddings by effectively incorporating topological features and EAGLE~\cite{wang2024effective} which proposes the factorized feature propagation (FFP) scheme for edge representations have shown up. Another line of research leverages Graph Neural Networks (GNNs).  In this scenario, one intuitive idea is to use the embeddings of the end nodes to form edge embeddings by some specific operations such as averaging, Hadamard product, concatenation, or by employing deep neural networks~\cite{kipf2016semi, velivckovic2017graph}. Methods like EHGNN~\cite{jo2021edge} can generate edge embeddings by transforming a graph's edges into the nodes of a hypergraph. Additionally, applying GNNs to the line graph derived from the original graph to generate embeddings for edges has also been explored~\cite{wang2020generic}. Furthermore, models such as EGNN~\cite{kim2019edge} and EGAT~\cite{gong2019exploiting}, although primarily designed for node classification, can directly produce edge embeddings. Moreover, some recent works~\cite{jiang2019censnet} aim to jointly learn node and edge embeddings to better capture the interplay between nodes and edges. Nevertheless, these models do not combine PLMs and GNNs to simultaneously process features on the edge, especially text.

\subsection{Textual Graph Representation Learning}
Representation learning for textual graphs seeks to construct high-quality node embeddings that capture the underlying textual and structural semantics, which has increasingly garnered considerable attention in academia.
Recent efforts towards this research direction fall into two major categories. One avenue of work integrates GNNs with PLM models. Early approaches, such as TextGNN~\cite{zhu2021textgnn}, AdsGNN~\cite{li2021adsgnn}, and GEAR~\cite{zhou2019gear}, integrate PLMs and GNNs in a cascaded fashion. More recent models resort to nested ways of combining PLMs and GNNs, exemplified by architectures like GLEM~\cite{zhao2022learning}, Graphformers~\cite{yang2021graphformers} and Edgeformers~\cite{jin2022edgeformers}. Beyond PLMs, another line of research focuses on developing TM-based text graph models. Notably, models like NetPLSA~\cite{mei2008topic} and RTM~\cite{chang2009relational} employ inference algorithms to optimize parameters, focusing on capturing graph structural information. A different stream of research involves models built primarily on deep neural networks. For instance, Adjacent-Encoder~\cite{zhang2020topic} and DBN~\cite{zhang2023topic} leverage latent themes to generate content for neighboring nodes, thereby enhancing the understanding of graph structures. Meanwhile, more sophisticated models, such as LANTM~\cite{wang2021layer}, GTNN~\cite{xie2021graph}, and GRTM~\cite{xie2021graph}, integrate GNNs into deep learning frameworks to facilitate more complex graph-based text analysis.

\section{Preliminaries}

\subsection{Notations and Terminology}

A TIN is defined as $\G=(\U, \V, \EDG, \DOC)$, where $\U$ and $\V$ symbolize a set of users and items, respectively, $\EDG$ contains interactions between users and items, and $\DOC$ consists of a collection of texts. Each interaction $e_{u,i}\in \EDG$ is endowed with a text ${T}_{e_{u,i}}\in \DOC$.

$\AM$ and $\dvec$ are used to denote the adjacency matrix and degree vector of a TIN $\G$,
respectively, wherein $\AM_{u,i}=\AM_{i,u}=1$ if $(u,i)\in \EDG$ and 0 otherwise, and $\dvec_{u}$ (resp. $\dvec_{i}$) is the degree of $u$ (resp. $i$).
The {\em unoriented} and {\em oriented} incidence matrices of $\G$ are symbolized by $\EM \in \mathbb{R}^{(|\U|+|\V|)\times |\EDG|}$ and $\BM \in \mathbb{R}^{(|\U|+|\V|)\times |\EDG|}$, respectively. For each interaction $e_{u,i}\in \EDG$ with user $u\in \U$ and item $i\in \V$, $\EM_{u,e_{u,i}}=\EM_{i,e_{u,i}}=1$ (resp. $\BM_{u,e_{u,i}}=1, \BM_{i,e_{u,i}}=-1$) but $\EM_{x,e_{u,i}}=0$ (resp. $\BM_{x,e_{u,i}}=0$) if $x\notin \{u,i\}$.
We use $\EUM\in \mathbb{R}^{|\EDG|\times |\U|}$ and $\EVM\in \mathbb{R}^{|\EDG|\times |\V|}$ to represent the columns of $\EM$ corresponding to users in $\U$ and items in $\V$, respectively. Throughout this paper, we refer to $e_{u,i}$ interchangeably as an interaction or an edge.

Based on the above definition,
we present the formal definition of the TIC problem studied in this paper in Definition~\ref{def:problem} as follows.
\begin{definition}[\bf Textual Interaction Classification]\label{def:problem}
Given a TIN $\G=(\U, \V, \EDG, \DOC)$ and the observed labels $\mathcal{Y}_{train}$ of a subset of interactions $\EDG_{train} \subset \EDG$, for each interaction $e_{u,i}\in \EDG\setminus \EDG_{train}$, the TIC task is to predict the class label of $e_{u,i}$ based on ${T}_{u,i}\in \DOC$, the graph topology surrounding $e_{u,i}$ in $\G$, and $\mathcal{Y}_{train}$ for $\EDG_{train}$.
\end{definition}

\subsection{Graph Neural Networks (GNNs)}\label{sec:GNN}

Most GNNs generally follow the {\em message-passing} paradigm~\cite{gilmer2017neural}, such as \textsf{GCN}~\cite{kipf2016semi}, \textsf{ChebyNet}~\cite{defferrard2016convolutional}, \textsf{SGC}~\cite{wu2019simplifying}, \textsf{APPNP}~\cite{gasteiger2018predict}, \textsf{GCNII}~\cite{chen2020simple}, and many others, where the node features are aggregated from the neighborhood along edges.
More concretely, in each $\ell$-th layer, the feature representation of $i$-th ($1\le i\le N$) node is updated via
\begin{small}
\begin{equation*}
\textstyle \HM^{(\ell+1)}_i = \sigma\left(\left(\sum_{j=1}^{N}{\PM_{i,j}\cdot \HM^{(\ell)}_j} \right)\cdot \WM^{(\ell)}\right),
\end{equation*}
\end{small}
where $\sigma(\cdot)$ denotes a nonlinear activate function, $\WM^{(\ell)}$ stands for the learnable parameters for feature transformation, and $\HM^{(0)}\in \mathbb{R}^{N\times d}$ is transformed from the node attributes. $\PM\in \mathbb{R}^{N\times N}$ is the aggregation matrix quantifying the strength of connections of nodes, which can be the normalized adjacency matrix, transition matrix, or the normalized Laplacian of the input graph. For instance, in \textsf{GCN}, $\PM$ is the normalized adjacency matrix $\NAM$, and \textsf{ChebyNet} uses Chebyshev polynomials of normalized graph Laplacian as $\PM$. 
\textsf{GCNII} adds a residual term $\alpha\HM^{(0)}$ to $\sum_{j=1}^{N}{\PM_{i,j}\cdot \HM^{(\ell)}_j}$ with $\PM=\NAM$, where $\alpha$ is a hyperparameter, and replaces $\WM^{(\ell)}$ by $\beta\IM+(1-\beta)\WM^{(\ell)}$.
In decoupled GNNs~\cite{dong2021equivalence}, e.g., \textsf{SGC}, \textsf{ANNP}, \textsf{JKNet}~\cite{xu2018representation} and \textsf{DAGNN}~\cite{liu2020towards}, the nonlinear activate function $\sigma(\cdot)$ is removed and $\WM^{(\ell)}$ is set to the identity matrix $\IM$ with merely one transformation for $\HM^{(0)}$.

\subsection{Transformer}\label{sec:transformer}

Transformer~\cite{vaswani2017attention} is a powerful deep learning architecture designed for sequence data, which has been widely adopted in PLMs, e.g., BERT~\cite{devlin2018bert}, in the form of multiple layers.
Inside each Transformer layer, a {\em multi-head attention} (MHA) mechanism is used to model the semantic relationships between tokens.
Let $\HM^{(0)} \in \mathbb{R}^{N \times d}$ be a sequence of $N$ token embeddings of the input text
and $\HM^{(\ell)} \in \mathbb{R}^{N \times d}$ be the output hidden states at the $\ell$-th Transformer layer.  
Particularly, at each $(\ell+1)$-th layer, the MHA is calculated by
\begin{equation}
    \textsf{MHA}(\HM^{(\ell)}) = \ccate_{h=1}^{H} \textsf{head}^{h}({\HM}^{(\ell)})
\end{equation}
where $\ccate$ stands for the horizontal concatenation of $H$ attention heads. Each attention head $\textsf{head}^{h}({\HM}^{(\ell)})$ is computed by
\begin{small}
\begin{equation}\label{eq:head}
\textstyle \textsf{head}^{h}({\HM}^{(\ell)}) = \textsf{softmax}\left(\frac{{\QM^{(h,\ell)}}{\KM^{(h,\ell)}}^{\top}}{\sqrt{d}} \right)\cdot \VM^{(h,\ell)},
\end{equation}
\end{small}
where $\QM^{(h,\ell)}$, $\KM^{(h,\ell)}$, and $\VM^{(h,\ell)}$ are known as queries, keys, and values, respectively, and are projected from ${\HM}^{(\ell-1)}$ with learnable weights $\WM^{(h,\ell)}_{Q}$, $\WM^{(h,\ell)}_{K}$, and $\WM^{(h,\ell)}_{V}$ as follows:
\begin{small}
\begin{align*}
\QM^{(h,\ell)} = {\HM}^{(\ell)}\WM^{(h,\ell)}_{Q},\ \KM^{(h,\ell)}={\HM}^{(\ell)}\WM^{(h,\ell)}_{K},\ \VM^{(h,\ell)}={\HM}^{(\ell)}\WM^{(h,\ell)}_{V}.
\end{align*}
\end{small}
The output hidden states $\HM^{(\ell+1)}$ is then obtained by applying a feed-forward network (\textsf{FFN}) to the MHA with a residual connection, followed by a layer normalization (\textsf{LaNorm}):
\begin{small}
\begin{equation}
\label{eq:transformer}
 \textstyle   \HM^{(\ell+1)} = \text{LaNorm}\left( \text{FFN}\left(\text{MHA}(\HM^{(\ell)})\right) + \HM^{(\ell)} \right).
\end{equation}
\end{small}

\vspace{-1ex}
\stitle{Connection to GNNs}
Let $\textstyle \SM = \textsf{softmax}\left({{\QM^{(h,\ell)}}{\KM^{(h,\ell)}}^{\top}}/{\sqrt{d}} \right)$. According to Eq.~\eqref{eq:head}, each attention head can be rewritten as
\begin{small}
\begin{equation*}
\textstyle \textsf{head}^{h}({\HM}^{(\ell)})_{i} = \left(\sum_{j=1}^{N}{ \SM_{i,j}\cdot {\HM}^{(\ell)}_{j}}\right)\cdot \WM^{(h,\ell)}_{V} .
\end{equation*}
\end{small}
Notice that $\sum_{j=1}^{N}{\SM_{i,j}}=1\ \forall{1\le i\le N}$ by the definition of the $\textsf{softmax}$ function. If we regard each text token as a node and $\HM^{(\ell)}$ as node features, the above equation for calculating the feature vector of $i$-th node in the $h$-th head is equivalent to aggregating transformed features of all nodes ${\HM}^{(\ell)}$ along edges with transition weights in the affinity graph $\SM$, followed by a linear transformation $\WM^{(h,\ell)}_{V}$.
Particularly, the affinity graph $\SM$ is learned from node features $\HM^{(\ell)}$.
As such, $\textsf{MHA}(\HM^{(\ell)})$ can be perceived as a concatenation of the node features obtained via message passing over $H$ learned affinity graphs.

\begin{figure*}[!t]
    \centering
    \includegraphics[width=0.95\textwidth]{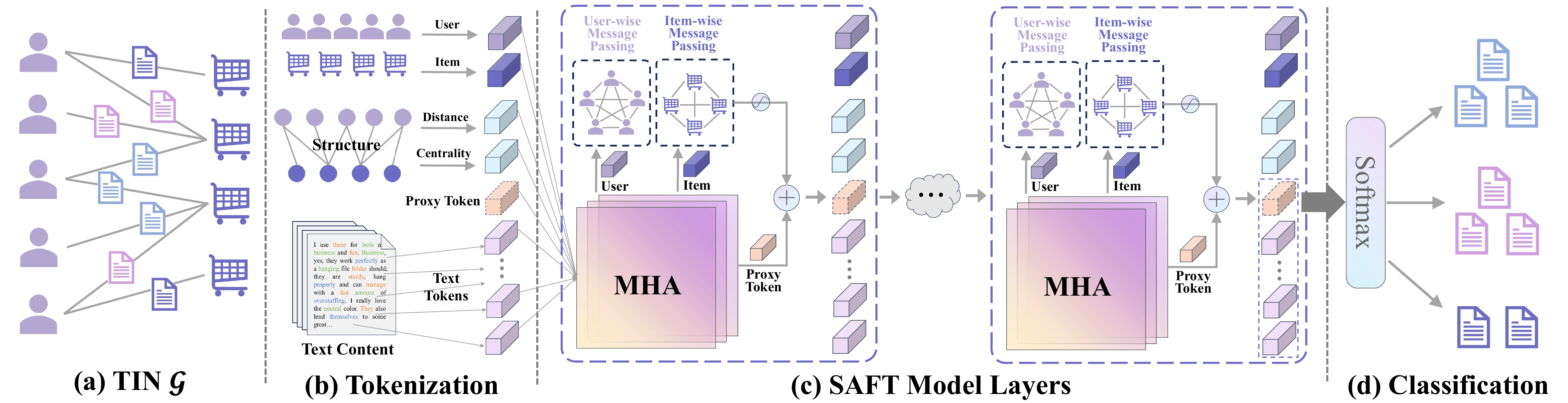} 
    \vspace{-2ex}
    \caption{The Overall Architecture of Our Proposed \algo. 
    }
    \label{fig:pipeline}
\vspace{0ex}
\end{figure*}

\section{Methodology}

In this section, we present our solution \algo for TIC. As depicted in Fig.~\ref{fig:pipeline}, \algo takes as input a TIN $\G$, and includes multiple Transformer layers alongside user- and item-wise message passing modules, where the former mainly seeks to encode contextualized semantics in the text $T_{e_{u,i}}$ of each interaction $e_{u,i}$ into token embeddings,
while the latter focuses on leveraging the correlations between textual interactions via users and items for feature learning. 
In addition to the text tokens of each textual interaction, the MHA operations in each layer are applied to tokens for its pertinent users and items, the structural embeddings (i.e., distance and centrality embeddings), as well as the proxy token. Particularly, the proxy token serves as an intermediary facilitating the message exchange between interactions and tokens, and Transformer layers. 

In succeeding subsections, we begin by introducing the module for textual interaction encoding in Section~\ref{sec:text-encoding}. After that, we elaborate on the user-/item-wise message passing schemes in Section~\ref{sec:user-item} and our fast theoretically-grounded algorithms for constructing distance and centrality embeddings from the TIN bipartite structure in Section~\ref{sec:struct-emb}. Section~\ref{sec:training} includes the label prediction, training objective, and our graph sampling strategies.

\subsection{Textual Interaction Encoding}\label{sec:text-encoding}
The textual interaction encoding in \algo mainly involves three stages: the token initialization at the first layer, embedding updates at the intermediate layers, and producing final representations at the last layer.

\subsubsection{\bf Token Initialization}
Let $N$ be the length of the text token sequence. Each textual interaction ${T}_{e_{u,i}}\in \DOC$ is then associated with a set of $N$ text tokens, which can be initially represented by $\TM^{(\textrm{t},0)}_{e_{u,i}}\in \mathbb{R}^{N \times d}$ and $d$ is the hidden dimension.
A proxy token $\TM^{(\textrm{p},0)}_{e_{u,i}} \in \mathbb{R}^{1\times d}$ is initialized as the mean of all the $N$ text tokens in $\TM^{(\textrm{t},0)}_{e_{u,i}}$. 
The tokens are then {\em vertically} concatenated as $\TM^{(\textrm{p},0)}_{e_{u,i}} \vcate \TM^{(\textrm{t},0)}_{e_{u,i}} \in \mathbb{R}^{(N+1) \times d}$,
which is later transformed into $\TM^{(0)}_{e_{u,i}}$ via MHA operations and a feed-forward network (\textsf{FFN}):
\begin{equation*}
\TM^{(1)}_{e_{u,i}} = \textsf{FFN}(\textsf{MHA}(\TM^{(\textrm{p},0)}_{e_{u,i}} \vcate \TM^{(\textrm{t},0)}_{e_{u,i}})).
\end{equation*}

In addition, for each textual interaction ${T}_{e_{u,i}}$, we create two tokens for user $u$ and item $i$, which are initialized based on their respective feature vectors $\XM^{(\textrm{u})}_{u}$ and $\XM^{(\textrm{i})}_{i}$ as follows:
\begin{equation}\label{eq:Xu-Xi}
{{\XM}^{(\textrm{u})}_{e_{u,i}} } = \textsf{LN}(\XM^{(\textrm{u})}_{u}),\quad {{\XM}^{(\textrm{i})}_{e_{u,i}}} = \textsf{LN}(\XM^{(\textrm{i})}_{i}),
\end{equation}
where $\textsf{LN}(\cdot)$ stands for a linear mapping. Particularly, $\XM^{(\textrm{u})}_{u}$ and $\XM^{(\textrm{i})}_{i}$ can come from the user and item attributes in the input TIN or be drawn from the Gaussian distribution.

To incorporate the topological patterns of interactions in $\G$, we construct {\em structural identity tokens} by
\begin{equation*}
{\ZM}_{e_{u,i}} = \textsf{LN}({\ZM}^{(\textrm{d})}_u) \vcate \textsf{LN}({\ZM}^{(\textrm{c})}_i),
\end{equation*}
where $\ZM^{(\textrm{d})}$ and $\ZM^{(\textrm{c})}$ represent the distance and centrality embeddings extracted from the TIN, respectively. We defer the rationale and algorithmic details for their generations to Section~\ref{sec:struct-emb}.

\subsubsection{\bf Embedding Updating}
Let $\TM^{(\ell)}_{e_{u,i}}$ be the output representation of interaction ${e_{u,i}}\in \EDG$ at the $\ell$-th ($\forall{1< \ell <L}$) layer in \algo model.
We first concatenate its text tokens, user, item tokens, and structural identity tokens as:
\begin{equation}
\textstyle {\HM}^{(\ell)}_{e_{u,i}} = {\TM^{(\ell-1)}_{e_{u,i}}} \vcate {{\XM}^{(\textrm{u})}_{e_{u,i}} } \vcate {{\XM}^{(\textrm{i})}_{e_{u,i}}}  \vcate {\ZM_{e_{u,i}}}.
\end{equation}
Subsequently, an \textsf{MHA} is applied to ${\HM}^{(\ell)}_{e_{u,i}}$ of each interaction ${e_{u,i}} \in \EDG$, which essentially updates all tokens within each interaction through the message passing between them, as pinpointed in Section~\ref{sec:transformer}. More specifically, the updated hidden states are as follows:
\begin{equation}\label{eq:update-H-MHA}
\begin{split}
\widetilde{\HM}^{(\ell)}_{e_{u,i}} 
= & \ \textsf{LaNorm}\left(\textsf{FFN}\left(\textsf{MHA}\left({\HM}^{(\ell)}_{e_{u,i}}\right)\right) + {\HM}^{(\ell)}_{e_{u,i}}\right),
\end{split}
\end{equation}
which can be represented by the vertical concatenation of updated token embeddings:
\begin{equation}\label{eq:update-H}
\textstyle \widetilde{\HM}^{(\ell)}_{e_{u,i}} = \widetilde{\TM}^{(\textrm{p},\ell-1)}_{e_{u,i}} \vcate \widetilde{\TM}^{(\textrm{t},\ell-1)}_{e_{u,i}} \vcate \widetilde{\XM}^{(\textrm{u})}_{e_{u,i}} \vcate \widetilde{\XM}^{(\textrm{i})}_{e_{u,i}} \vcate \widetilde{\ZM}_{e_{u,i}}.
\end{equation}

Intuitively, through the above token-token message passing, new token embeddings $\widetilde{\XM}^{(\textrm{u})}_{e_{u,i}}$ and $\widetilde{\XM}^{(\textrm{i})}_{e_{u,i}}$ for user $u$ and item $i$ assimilate the textual and topological semantics from ${\TM}^{(\textrm{t},\ell-1)}_{e_{u,i}}$ and ${\ZM}_{e_{u,i}}$.

Based thereon, \algo proceeds to interaction-level message passing, which aims to aggregate features from neighboring users and items along their connections in $\G$. Denote by $\widehat{\XM}^{(\textrm{u})}_{e_{u,i}}$ and $\widehat{\XM}^{(\textrm{i})}_{e_{u,i}}$ the normalized representations of $e_{u,i}$ output by the user-wise and item-wise message passing modules (see Section~\ref{sec:user-item}), respectively. The proxy token embedding is then updated by injecting the newly aggregated user and item features in $\widehat{\XM}^{(\textrm{u})}_{e_{u,i}}$ and $\widehat{\XM}^{(\textrm{i})}_{e_{u,i}}$:
\begin{small}
\begin{equation*}
\textstyle \widehat{\TM}^{(\textrm{p},\ell-1)}_{e_{u,i}} = \textsf{ReLU}\left( \textsf{LN}\left( \widetilde{\TM}^{(\textrm{p},\ell-1)}_{e_{u,i}} \vcate \widehat{\XM}^{(\textrm{u})}_{e_{u,i}} \vcate \widehat{\XM}^{(\textrm{i})}_{e_{u,i}} \right)\right).
\end{equation*}
\end{small}
Accordingly, the output representation $\TM^{(\ell)}_{e_{u,i}}$ for $e_{u,i}$ at this layer is updated by $\textstyle \TM^{(\ell)}_{e_{u,i}} = {\widehat{\TM}^{(\textrm{p},\ell-1)}_{e_{u,i}}} \vcate {\widetilde{\TM}^{(\textrm{t},\ell-1)}_{e_{u,i}}}$.

\subsubsection{\bf Final Representations}
Instead of directly taking the average of token embeddings in ${\TM}^{(L)}_{e_{u,i}}$ output by the final layer of \algo, i.e., $L$-th layer, as the feature representation of interaction $e_{u,i}$ for TIC, we resort to a pooling of the last two layers~\cite{li2020sentence}. More precisely, the final representation for each $e_{u,i}\in \EDG$ is computed by averaging the outputs of the final and penultimate layers:
\begin{equation}\label{eq:final-emd}
\TM_{e_{u,i}} = \textsf{mean}(\overline{\TM}^{(L)}_{e_{u,i}}, \overline{\TM}^{(L-1)}_{e_{u,i}}),
\end{equation}
where $\overline{\TM}^{(L)}_{e_{u,i}}$ and $\overline{\TM}^{(L-1)}_{e_{u,i}}$ stand for the averages of their respective token embeddings.

\subsection{User-wise and Item-wise Message Passing}\label{sec:user-item}

At each $\ell$-th ($\ell \in (1,L]$) layer in \algo, the user-wise (resp. item-wise) message passing module focuses on transforming the user (resp. item) token embedding $\widetilde{\XM}^{(\textrm{u})}_{e_{u,i}}$ (resp. $\widetilde{\XM}^{(\textrm{i})}_{e_{u,i}}$) of each edge $e_{u,i}\in \EDG$ obtained in Eq.~\eqref{eq:update-H} into augmented user features $\widehat{\XM}^{(\textrm{u})}_{e_{u,i}}$ (resp. $\widehat{\XM}^{(\textrm{i})}_{e_{u,i}}$).
To aggregate user and item features of other edges, an idea is to apply the MHA mechanism as in the token-to-token message passing in Eq.~\eqref{eq:update-H-MHA} to all interactions. However, this approach yields a prohibitively high computational complexity of $O(|\EDG|^2)$ and falls short of the exploitation of the graph topology of $\G$. In what follows, we propose two graph-based workarounds to fulfill the linear-complexity message passing between interactions via users and items.

\subsubsection{\bf LGA-based Method}\label{subsec:messagePassing}
Our first method is to build two line graphs with interactions in $\EDG$ as nodes based on their associations with users and items in $\G$, respectively, and then follow the message passing rules in graphs.

Recall that $\EUM \in \mathbb{R}^{|\EDG| \times |\U|}$ and $\EVM \in \mathbb{R}^{|\EDG| \times |\V|}$ denote the incidence matrices pertaining to users and items, respectively, wherein $\EUM_{e_{u,i},v} = 1$ (resp. $\EVM_{e_{u,i},\ell} = 1$) if $u=v$ (resp. $i=\ell$) and $0$ otherwise.
Accordingly, the two line graphs can be modeled by their respective transition matrices defined by
\begin{small}
\begin{equation}\label{eq:PUV}
\PUM = \EUM \DUM^{-1} \EUM^{\top}\ \text{and}\ \PVM = \EVM \DVM^{-1} \EVM^{\top},
\end{equation}
\end{small}
where $\dvec^{(\textrm{u})}_u = \dvec_u+1\ \forall{u\in \U}$ (resp. $\dvec^{(\textrm{i})}_i = \dvec_i+1\ \forall{i\in \V}$).
In particular, $\PUM_{e_{u,i},e_{v,\ell}}=\frac{1}{\dvec_u+1}$ if $u=v$, meaning that the two interactions $e_{u,i},e_{u,\ell}$ are connected to each other via a common user $u$, and otherwise $\PUM_{e_{u,i},e_{v,\ell}}=0$. 
Analogously, $\PVM_{e_{u,i},e_{v,\ell}}=\frac{1}{\dvec_i+1}$ if $i=\ell$ and $0$ otherwise.

With these two line graphs, we iteratively update the user and item features as 
remarked in Section~\ref{sec:GNN}. More specifically, at $r$-th iteration, the user and item embeddings $\UM^{(\textrm{u},r)}$ and $\UM^{(\textrm{i},r)}$ ($1\le r\le R$) are calculated by
\begin{small}
\begin{equation}\label{eq:GNN-UV}
\begin{split}
\UM^{(\textrm{u}, r)} = \PUM \UM^{(\textrm{u}, r-1)} + \delta\cdot {\UM^{(\textrm{u}, 0)}},\\ \UM^{(\textrm{i}, r)} = \PVM \UM^{(\textrm{i}, r-1)} + \delta\cdot {\UM^{(\textrm{i}, 0)}},
\end{split}
\end{equation}
\end{small}
where $\UM^{(\textrm{u}, 0)}$ (resp. $\UM^{(\textrm{i}, 0)}$) signifies the initial user (resp. item) embeddings and $\delta$ is a hyper-parameter for adding residual connections. In lieu of simply employing the updated user and item token embedding $\widetilde{\XM}^{(\textrm{u})}_{e_{u,i}}$ (resp. $\widetilde{\XM}^{(\textrm{i})}_{e_{u,i}}$) from Eq.~\eqref{eq:update-H} as $\UM^{(\textrm{u}, 0)}_{e_{u,i}}$ and $\UM^{(\textrm{i}, 0)}_{e_{u,i}}$, we inject the original user and item features in Eq.~\eqref{eq:Xu-Xi} as residual connections from the first layer, which leads to 
\begin{equation}\label{eq:init-U}
\begin{gathered}
{\UM^{(\textrm{u}, 0)}}_{e_{u,i}} =  \widetilde{\XM}^{(u)}_{e_{u,i}} + \lambda\cdot \XM^{(u)}_{e_{u,i}},\ {\UM^{(\textrm{i}, 0)}}_{e_{u,i}} =  \widetilde{\XM}^{(i)}_{e_{u,i}} + \lambda\cdot \XM^{(i)}_{e_{u,i}}, \\
\end{gathered}
\end{equation}
where $\lambda$ is a hyperparameter.

After $R$ iterations of feature aggregations, \algo generates the new user and item features as follows: 
\begin{small}
\begin{equation}\label{eq:new-X}
\begin{split}
\widehat{\XM}^{(\textrm{u})}_{e_{u,i}} = \textsf{LaNorm}\left(\textsf{LN}\left(\UM^{(\textrm{u}, R)}_{e_{u,i}}\right)\right),\ \widehat{\XM}^{(\textrm{i})}_{e_{u,i}} = \textsf{LaNorm}\left(\textsf{LN}\left(\UM^{(\textrm{i}, R)}_{e_{u,i}}\right)\right).
\end{split}
\end{equation}
\end{small}

\stitle{Analysis} 
Let $\textsf{\textnormal{ssoftmax}}(\cdot)$ be a sparse softmax function, which calculates the softmax merely for non-zero entries in the input matrix.
If we replace $\QM^{(h,\ell)}$, $\KM^{(h,\ell)}$ in Eq.~\eqref{eq:head} by $\EUM$, $\VM^{(h,\ell)}$ by $\UM^{(\textrm{u}, r-1)}$, and the softmax function by $\textsf{\textnormal{ssoftmax}}(\cdot)$, Lemma~\ref{lem:softmax-P}\footnote{All proofs appear in Appendix~\ref{sec:proof}.} implies that the resulting attention head is equal to $\PUM \UM^{(\textrm{u}, r-1)}$. A similar result can be derived for $\PVM \UM^{(\textrm{i}, r-1)}$. These observations reveal that our above feature aggregation operation in Eq.~\eqref{eq:GNN-UV} is basically an attention mechanism using the incidence matrix $\EUM$ or $\EVM$ as queries and keys, referred to as {\em line graph attention} (LGA).
\begin{lemma}\label{lem:softmax-P}
$\textstyle \PUM=\textsf{\textnormal{ssoftmax}}\left({\EUM} {\EUM}^{\top}/\sqrt{|\U|}\right)$ and $\textstyle \PVM=\textsf{\textnormal{ssoftmax}}\left({\EVM} {\EVM}^{\top}/\sqrt{|\V|}\right)$ hold.
\end{lemma}

\subsubsection{\bf GAU-based Method}\label{sec:GAU}
Inspired by~\cite{hua2022transformer}, another substitute for applying \textsf{MHA} operations over interactions for message passing is the {\em gated attention unit} (GAU), which is built upon the {\em gated linear unit}~\cite{shazeer2020glu}. Although it reduces multiple heads in the vanilla Transformer to a single head whilst achieving matching or even superior performance, it still requires learning the attention operator (i.e., the affinity graph), which is intolerable for interactions in $\EDG$.
Instead, we utilize $\PUM$ and $\PVM$ as the attention operators in this GAU module.
In turn, the user embeddings $\UM^{(\textrm{u}, r)}$ and item embeddings $\UM^{(\textrm{i}, r)}$ ($1\le r\le R$) are iteratively updated as follows:
\begin{small}
\begin{equation}\label{eq:GAU-UV}
\begin{split}
\UM^{(\textrm{u}, r)} = \textsf{LN}\left(\GaM^{(\textrm{u})} \odot \left(\PUM\UM^{(\textrm{u}, r-1)} + \delta\cdot \DeM^{(\textrm{u})})\right)\right),\\
\UM^{(\textrm{i}, r)} = \textsf{LN}\left(\GaM^{(\textrm{i})} \odot \left(\PVM\UM^{(\textrm{i}, r-1)} + \delta\cdot \DeM^{(\textrm{i})})\right)\right),
\end{split}
\end{equation}
\end{small}
whereas the initial user/item embeddings and final user/item features are constructed the same as Eq.~\eqref{eq:init-U} and Eq.~\eqref{eq:new-X}, respectively.
$\GaM^{(\textrm{u})}, \DeM^{(\textrm{u})}$ and $\GaM^{(\textrm{i})}, \DeM^{(\textrm{i})}$ are transformed from the initial user and item embeddings $\UM^{(\textrm{u}, 0)}$ and ${\UM^{(\textrm{i}, 0)}}$ by
\begin{small}
\begin{equation*}
\begin{split}
\GaM^{(\textrm{u})} = \textsf{SiLU}\left(\textsf{LN}\left( {\UM^{(\textrm{u}, 0)}} \right)\right), \DeM^{(\textrm{u})} = \textsf{SiLU}\left(\textsf{LN}\left({\UM^{(\textrm{u}, 0)}} \right)\right),\\
\GaM^{(\textrm{i})} = \textsf{SiLU}\left(\textsf{LN}\left( {\UM^{(\textrm{i}, 0)}} \right)\right), \DeM^{(\textrm{i})} = \textsf{SiLU}\left(\textsf{LN}\left({\UM^{(\textrm{i}, 0)}} \right)\right),
\end{split}
\end{equation*}
\end{small}
where \textsf{SiLU} represents the {\em sigmoid linear unit} activation function.

\subsubsection{\bf Complexity Analysis}
Notice that the major computational overhead incurred by these two methods lies in the operations in Eq.~\eqref{eq:GNN-UV} and Eq.~\eqref{eq:GAU-UV}.
The explicit constructions of $\PUM$ and $\PVM$ need $O(\sum_{u\in \U}{\dvec_u^2})$ and $O(\sum_{i\in \V}{\dvec_i^2})$ time costs, which are $O(|\U|^2)$ and $O(|\V|^2)$ in the worst case, respectively.
Note that such issues can be circumvented by reordering the matrix multiplications in Eq.~\eqref{eq:GNN-UV} and Eq.~\eqref{eq:GAU-UV} as $\textstyle \EUM \DUM^{-1} \cdot \left(\EUM^{\top}\UM^{(r-1)}\right)$ and $\textstyle \EVM \DVM^{-1} \cdot \left(\EVM^{\top}\VM^{(r-1)}\right)$, whereby the computational costs are reduced to $O(|\EDG|\cdot d)$. Since the linear transforms in Sections~\ref{subsec:messagePassing} and ~\ref{sec:GAU} take $O(|\EDG|\cdot d^2)$ time, and other matrix additions and Hadamard products require $O(|\EDG|\cdot d)$ time, the overall time complexity is then bounded by $O(|\EDG|\cdot d^2)$.

\begin{figure}[t!]
    \centering
    \includegraphics[width=0.45\textwidth]{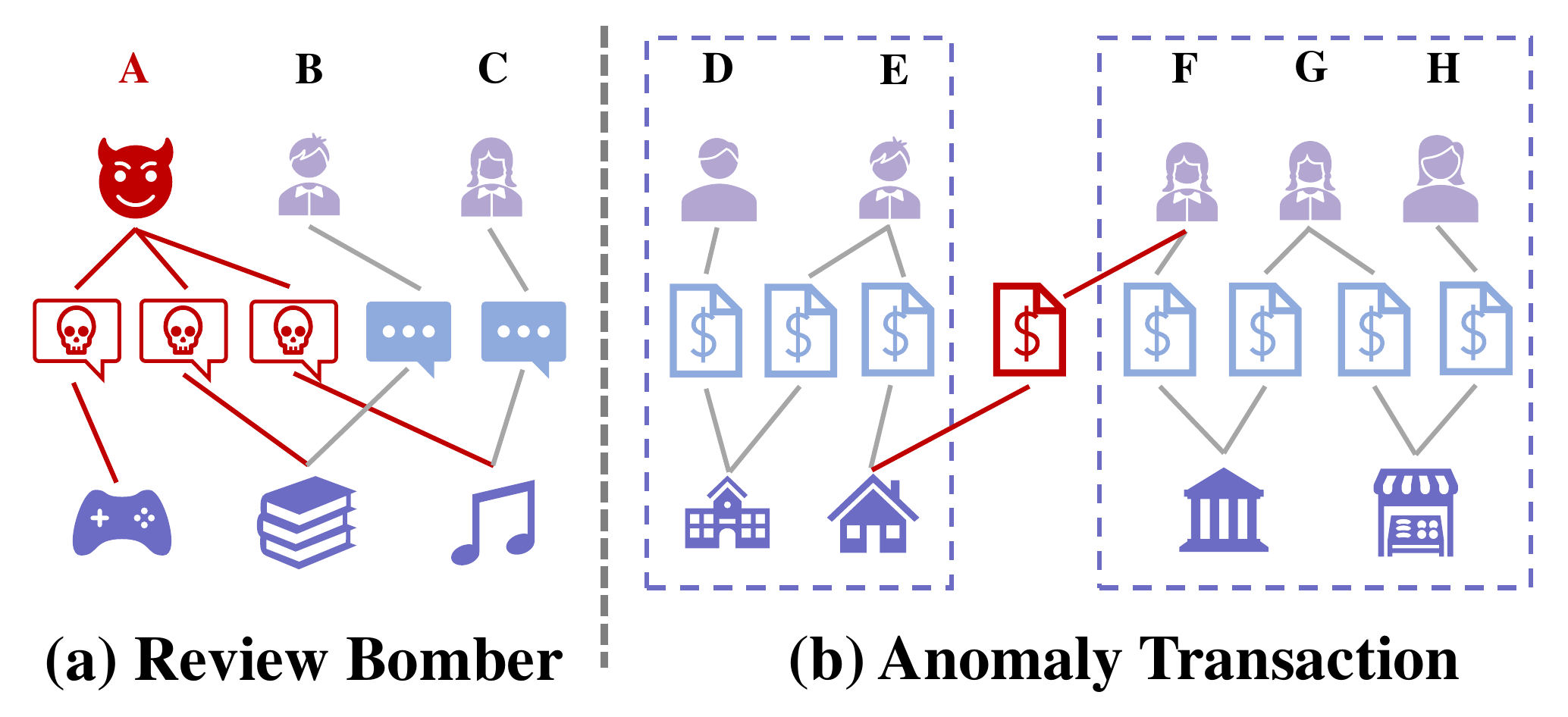} 
    \vspace{-3ex}
    \caption{Local and Global Structural Patterns in TIN $\G$.}
    \label{fig:examples}
    \vspace{-2ex}
\end{figure}

\subsection{Structural Encoding for Interactions}\label{sec:struct-emb}

The local and global structural patterns pertinent to interactions underlying the TIN $\G$ present rich semantics conducive to TIC.
To illustrate, Fig.~\ref{fig:examples} shows two example TINs.
Fig.~\ref{fig:examples}(a) exemplifies malicious reviews (symbolized by skulls) from a "Review Bomber", i.e., customer A, to a set of related goods. These reviews are locally proximal to each other on the interaction graph as they are from the same user and involve correlated products. With such local patterns, it is more likely to classify them as malicious correctly.
Fig.~\ref{fig:examples}(b) displays the transactions between credit card users and merchants, wherein \{D,E\} and \{F,G,H\} represent two communities of users in different regions. Observe that these distant communities are bridged via a transaction from user F (highlighted in red). Intuitively, such a transaction is suspicious and might be caused by credit card fraudulent activity. 
This observation implies the usefulness of the structural roles of transactions in the global graph for detecting anomaly transactions.
Inspired by these insights, we propose to encode the local and global structural information into {\em distance embeddings} and {\em centrality embeddings} as follows.

\subsubsection{\bf Distance Embedding}
To facilitate the encoding of the local structures surrounding interactions, we first construct a line graph $\widetilde{\G}$ of $\G$, in which each interaction is regarded as a node and interactions are connected via edges in $\widetilde{\G}$ if they are associated with the same users or items in $\G$. Accordingly, the weighted adjacency matrix $\PM$ of $\widetilde{\G}$ is defined by $\EM^{\top}\frac{\DM^{-1}}{2}\EM$, wherein $\PM_{e_{u,i}, e_{u,j}}$ signifies the probability of a random walk jumping from interaction $e_{u,i}$ to interaction $e_{u,j}$ of user $u$ using $u$ as the intermediary.

Let $\UM\boldsymbol{\Sigma}{\VM}^{\top}$ be the top-$k$ SVD of $\DM^{-\frac{1}{2}}{\EM}$. We calculate the distance embeddings $\ZM^{(\textrm{d})}\in \mathbb{R}^{|\EDG|\times k}$ of interactions in $\EDG$ by
\begin{small}
\begin{equation}\label{eq:dist-emb}
\textstyle \ZM^{(\textrm{d})} = \VM \sqrt{\frac{1}{\IM - {\boldsymbol{\Sigma}}^2/2}}.
\end{equation}
\end{small}
\begin{theorem}\label{lem:Lap-Inv}
When $k=|\U|+|\V|$, $\forall{e_{u,i}, e_{u,j}\in \EDG}$, $ \|\ZM^{(\textrm{d})}_{e_{u,i}} -\ZM^{(\textrm{d})}_{e_{u,j}}\|^2=RD(e_{u,i},e_{u,j})$.
\end{theorem}

Theorem~\ref{lem:Lap-Inv} implies that our distance embeddings $\ZM^{(\textrm{d})}$ capture the prominent {\em resistance distance}~\cite{klein1993resistance,yang2023efficient} between nodes in the graph $\widetilde{\G}$. Let $\widetilde{\LM}$ be the Laplacian matrix of $\widetilde{\G}$ and $\widetilde{\LM}^{\dagger}$ be its pseudo-inverse. In mathematical terms, the resistance distance $RD(e_{u,i}, e_{u,j})$ of interactions $e_{u,i}, e_{u,j}$ on $\widetilde{\G}$ is defined by
\begin{equation*}
RD(e_{u,i}, e_{u,j}) = \widetilde{\LM}^{\dagger}_{e_{u,i},e_{u,i}} + \widetilde{\LM}^{\dagger}_{e_{u,j},e_{u,j}} -2 \widetilde{\LM}^{\dagger}_{e_{u,i},e_{u,j}}.
\end{equation*}
According to~\cite{lovasz1993random}, a large $RD(e_{u,i}, e_{u,j})$ indicates more expected steps needed for a random walk originating from $e_{u,i}$ visits $e_{u,j}$ and then goes back to $e_{u,i}$. Intuitively, if two interactions $e_{u,i}, e_{u,j}$ are distant from each other on $\widetilde{\G}$, their resistance distance $RD(e_{u,i}, e_{u,j})$ should be large, thereby engendering dissimilar distance embeddings $\ZM^{(\textrm{d})}_{e_{u,i}}$ and $\ZM^{(\textrm{d})}_{e_{u,j}}$.

Furthermore, the following theorem states that our distance embeddings $\ZM^{(\textrm{d})}$ preserve the multi-hop topological proximity (i.e., {\em Katz index}~\cite{katz1953new}) between nodes in $\widetilde{G}$ (i.e., interactions in $\G$).
\begin{theorem}\label{lem:Katz}
When $k=|\U|+|\V|$, we have $\ZM^{(\textrm{d})}_{e_{u,i}} \cdot {\ZM^{(\textrm{d})}}_{e_{u,j}}^{\top} = \sum_{\ell=0}^{\infty}{\alpha^\ell\PM^\ell}_{e_{u,i},e_{u,j}}$,
where $\sum_{k=0}^{\infty}{\alpha^k\PM^k}_{e_{u,i},e_{u,j}}$ stands for the Katz index of $e_{u,j}$ w.r.t. $e_{u,i}$ over $\widetilde{\G}$ with weight $\alpha=1$.
\end{theorem}

Note that in practice, the dimension $k$ of distance embeddings is set to a small integer, e.g., $128$, instead of $|\U|+|\V|$, for both computation and space efficiency.

\subsubsection{\bf Centrality Embedding}
Next, we seek to encode the centrality/importance of each interaction in $\G$ from the perspective of the entire graph into centrality embeddings $\ZM^{(\textrm{c})}\in \mathbb{R}^{|\EDG|\times k}$.
The {\em spanning centrality}~\cite{mavroforakis2015spanning,zhang2023efficient} $s(e_{u,i})$ is adopted to quantify the importance of each interaction $e_{u,i}$ in $\G$, which is defined by
\begin{equation}
s(e_{u,i})={|\tau(\G)_{e_{u,i}}|}/{|\tau(\G)|},
\end{equation}
where $\tau(\G)$ denotes the set of spanning trees of $\G$ and $\tau(\G)_{e_{u,i}}$ contains the spanning trees containing $e_{u,i}$. In essence, $s(e_{u,i})$ equals the fraction of spanning trees of $\G$ containing $e_{u,i}$. Intuitively, a higher $s(e_{u,i})$ indicates that $e_{u,i}$ participates in more spanning trees of $\G$, and, thus, is more important to $\G$.

Let $\PhiM\LaM\PsiM^{\top}$ be the top-$k$ SVD of the oriented incidence matrix $\BM$ of $\G$. Our centrality embeddings are computed via $\ZM^{(\textrm{c})} = \PsiM$.
\begin{theorem}\label{lem:centrality}
When $k=|\U|+|\V|$, $s(e_{u,i})=\|\PsiM_{e_{u,i}}\|^2_2$ $\forall{e_{u,i}\in \EDG}$.
\end{theorem}
Theorem~\ref{lem:centrality} shows that $\ZM^{(\textrm{c})}$ preserves the spanning centrality of each interaction in $\G$.

\subsubsection{\bf Complexity Analysis}
The main computational overhead in forming the embeddings $\ZM^{(\textrm{d})}$ and $\ZM^{(\textrm{c})}$ stems from computing the top-$k$ singular value decompositions (SVDs) of large sparse matrices. For $\ZM^{(\textrm{d})}$, as defined in Eq.~\eqref{eq:dist-emb}, we require the top-$k$ SVD of the matrix $\DM^{-\frac{1}{2}}\EM$, where $\EM \in \mathbb{R}^{(|\U|+|\V|)\times |\EDG|}$ is the incidence matrix and $\DM$ is the degree matrix. Similarly, constructing $\ZM^{(\textrm{c})}$ involves computing the top-$k$ SVD of the oriented incidence matrix $\BM \in \mathbb{R}^{(|\U|+|\V|)\times |\EDG|}$. Directly computing these SVDs incurs a time complexity of $O((|\U|+|\V|)^{2} \cdot |\EDG|)$ or $O((|\U|+|\V|) \cdot |\EDG|^{2})$, which is impractical for large graphs. To alleviate this, we employ the Randomized SVD to leverage the sparsity of $\EM$ and $\BM$ and resolve sign ambiguity, whose time complexity of this algorithm is $O(|\EDG|\cdot k^{2})$. Additional operations, such as scaling and element-wise transformations, require $O(|\EDG| \cdot k)$ time. Therefore, the overall time complexity for constructing both $\ZM^{(\textrm{d})}$ and $\ZM^{(\textrm{c})}$ is bounded by $O(|\EDG|\cdot k^{2})$, ensuring scalability for large graphs when $k$ is modest.

\subsection{Model Training}\label{sec:training}

\subsubsection{\bf Training Objective}
Let $\YM\in \mathbb{R}^{|\EDG_{train}|\times K}$ denote the label predictions of interactions in $\EDG_{train}$, where $K$ is the number of class labels.
For each edge $e_{u,i}\in \EDG_{train}$, $\YM_{e_{u,i}}$ can be obtained by a linear transformation of its representation ${\TM}_{e_{u,i}}$ computed at Eq.~\eqref{eq:final-emd}, followed by a softmax:
\begin{equation*}
\YM_{e_{u,i}} = \textsf{softmax}\left(\textsf{LN}\left({\TM}_{e_{u,i}}\right)\right).
\end{equation*}

Following common practice, we adopt the cross-entropy loss with ground-truth labels in $\mathcal{Y}_{train}$ for the training set $\EDG_{train}$ to train our \algo model in a supervised fashion:
\begin{small}
\begin{equation*}
- \sum_{e_{u,i}\in \EDG_{train}}\sum_{k=1}^{K}{ \YM^{\ast}_{e_{u,i},k} \log{({\YM}_{e_{u,i},k})} + (1 - \YM^{\ast}_{e_{u,i},k}) \log (1 - {\YM}_{e_{u,i},k})},
\end{equation*}
\end{small}
where $\YM^{\ast}\in \mathbb{R}^{|\EDG_{train}|\times K}$ contains the group-truth labels of interactions in $\EDG_{train}$. Particularly, $\YM^{\ast}_{e_{u,i},k}=1$ if $e_{u,i}$ truly belongs to the $k$-th class, and $0$ otherwise.

\begin{table}[!t] 
\renewcommand{\arraystretch}{0.8}
\caption{Dataset Statistics}
\label{tab:dataset_statistics}
\vspace{-3ex}
\begin{center}
\begin{small}
\begin{tabular}{ l|c|c|c|c } 
\hline
{\bf Dataset} & {\bf $|\U|$} & {\bf $|\V|$} & {\bf $|\EDG|$} & {\bf $K$ } \\
\hline
Goodreads-Children & 10,521 & 1,479 & 40,762 & 6\\ 
Amazon-Apps & 4,390 & 5,610 & 51,073 & 5\\ 
Amazon-Movie & 4,431 & 1,819 & 61,216 & 5\\
Goodreads-Crime & 7,009 & 1,241 & 62,774 & 6\\ \hline 
Goodreads-Poetry & 47,400 & 36,412 & 154,555 & 6\\ 
Google-Vermont & 12,655 & 5,040 & 178,168 & 5\\ 
Google-Hawaii & 39,215 & 10,170 &  710,948& 5\\ 
Amazon-Products & 101,498 & 27,965 & 800,144 & 5\\ 
\hline
\end{tabular}
\end{small}
\end{center}
\vspace{-2ex}
\end{table}

\subsubsection{\bf Graph Sampling} \label{sec:graphSampling}
Recall that the feature aggregation operations in Eq.~\eqref{eq:GNN-UV} and Eq.~\eqref{eq:GAU-UV} involve the entire graph topology $\PUM$ and $\PVM$, which is impractical for large TINs.
To mitigate this issue, we resort to sampling $b$ adjacent interactions of $e_{u,i}\in \EDG$ from $\{e_{u,\ell}|\ell \in \N(u)\}$ and $\{e_{v,i}|v \in \N(i)\}$ to form $\Ss_{u}$ and $\Ss_{i}$ for user-wise and item-wise feature aggregations, respectively, where $\N(u)$ (resp. $\N(i)$) symbolizes the set of neighboring items (resp. items) of user $u$ (resp. item $i$).
For interactions incident to the same user $u$ (resp. item $i$), they share the same $\Ss_{u}$ (resp. $\Ss_{i}$) set.
Note that when $|\N(u)|\le b$ (resp. $|\N(i)|\le b$), $\Ss_{u}=\{\ZM^{(\textrm{d})}_{e_{u,\ell}}|\ell \in \N(u)\}$ (resp. $\Ss_{i}=\{\ZM^{(\textrm{d})}_{e_{v,i}}|v \in \N(i)\}$).

\header
{\bf Distance-based sampling} picks $b$ interactions based on their topological distances to the target $e_{u,i}$ from the perspectives of user $u$ and item $i$, respectively. Theorem~\ref{lem:Katz} states that the dot product of the distance embeddings $\ZM^{(\textrm{d})}_{e_{u,i}}$ and $\ZM^{(\textrm{d})}_{e_{u,\ell}}$ of two edges $e_{u,i}$ and $e_{u,\ell}$ quantifies their proximity over the graph. Accordingly, we sample user $u$'s adjacent edge $e_{u,x}$ and item $i$'s adjacent edge $e_{v,i}$ with a probability proportional to $\textstyle \frac{\ZM^{(\textrm{d})}_{e_{u,i}}\cdot \ZM^{(\textrm{d})}_{e_{u,x}}}{\sum_{\ell\in \N(u)}{\ZM^{(\textrm{d})}_{e_{u,i}}\cdot \ZM^{(\textrm{d})}_{e_{u,\ell}}}}\ \text{and}\ \frac{\ZM^{(\textrm{d})}_{e_{u,i}}\cdot \ZM^{(\textrm{d})}_{e_{v,i}}}{\sum_{\mu\in \N(i)}{\ZM^{(\textrm{d})}_{e_{u,i}}\cdot \ZM^{(\textrm{d})}_{e_{\mu,i}}}}$,
and then add them to $\Ss_{u}$ and $\Ss_{i}$, respectively. 

\begin{table*}[!t] 
\centering
\renewcommand{\arraystretch}{0.9}
\caption{TIC performance on small datasets. The best results are bolded, while the best baselines are \underline{underlined}. 
}
\label{tab:small_datasets}
\vspace{-2ex}
\begin{small}
\addtolength{\tabcolsep}{-0.25em}
\resizebox{0.9\textwidth}{!}{
\begin{tabular}{ c | c  c  | c  c  | c  c   | c  c  }
\hline
\multirow{2}{*}{\textbf{Model}} & \multicolumn{2}{ c |}{\textbf{Goodreads-Children}} & \multicolumn{2}{ c |}{\textbf{Amazon-Apps}} & \multicolumn{2}{ c |}{\textbf{Amazon-Movie}} & \multicolumn{2}{ c }{\textbf{Goodreads-Crime}}  \\ 
\cline{2-9}
& {Macro-F1} \textuparrow & {Micro-F1} \textuparrow & {Macro-F1} \textuparrow & {Micro-F1} \textuparrow & {Macro-F1} \textuparrow & {Micro-F1} \textuparrow & {Macro-F1} \textuparrow & {Micro-F1} \textuparrow \\
\hline

TF-IDF  &17.67&38.21&21.33&35.61&20.26&27.87&17.29&30.64\\
TF-IDF+NODES &16.40&43.05&17.86&38.96&15.96&34.10&12.44&35.17\\ \hline
AttrE2vec~\cite{bielak2022attre2vec} &11.55&27.58&13.92&34.17&10.98&29.58&5.93&21.62\\
TER+AER (PPR)~\cite{wang2023efficient} & \underline{36.85} &49.28&34.14&56.45&33.20&42.14&33.50&49.34\\
TER+AER (HKPR)~\cite{wang2023efficient} &35.76&49.96&34.26&56.20&35.28&43.56&35.76& 49.96 \\
EAGLE~\cite{wang2024effective} &29.86&48.80&25.23&54.14&24.88&41.55&28.16&46.72\\ \hline

BERT~\cite{devlin2018bert}  &29.86&48.80&25.23&54.14&24.88&41.55&28.16&46.72\\

BERT+NODES &35.60& \underline{50.36} &43.85& 57.13 &40.87&44.61&35.66&47.23\\

BERT+GraphSAGE~\cite{hamilton2017inductive} &36.25&49.93&43.89&56.14& \underline{41.02} & 45.76 &35.68&46.38\\
Deberta~\cite{he2020deberta} + GCN & 35.65 & 49.35 & 43.74 & 56.89 & 39.85 & 45.47 & 35.66 & 48.90 \\
Sentence-BERT~\cite{reimers2019sentence} + GCN & 35.88 & 49.50 & 44.08 & \underline{57.42} & 40.03 & 45.62 & 35.91 & 49.50 \\
LLaMA-2 (7B)~\cite{touvron2023llama} + GCN & 34.93 & 48.41 & 43.03 & 55.90 & 38.92 & 44.68 & 34.89 & 48.00 \\
\hline
GraphFormers~\cite{yang2021graphformers} &36.04&49.97& \underline{44.44} &57.03&39.86&44.05& \underline{36.51} &47.04\\ 
GIANT~\cite{chien2021node} & 35.79 & 49.62 & 43.87 & 56.95 & 39.98 & 45.64 & 35.72 & 49.10 \\
GLEM-LLM~\cite{chen2024exploring} & 35.89 & 49.76 & 44.12 &57.02 &40.13&45.87&35.89&49.50 \\
GLEM-GNN~\cite{chen2024exploring} & 36.12 & 50.01 & 44.25 & 57.25 & 40.32 &\underline{45.92}&36.04& \underline{50.00} \\
Edgeformers~\cite{jin2023edgeformers} &36.79&48.66&43.60&56.35&40.91&45.46&35.50&47.14\\ \hline
    
\algognn &36.95&\textbf{51.68}& \textbf{45.92} & \textbf{61.06} & \textbf{41.24} &46.41 &\textbf{36.57} &50.21\\
Improv. &+0.10&+1.32&+1.48&+3.64&+0.22&+0.49&+0.06&+0.21\\
\algogau &\textbf{36.98} &51.17 & 45.07 & 59.06 &41.08& \textbf{46.57} &36.51&\textbf{50.33} \\
Improv. &+0.13&+0.81&+0.63&+1.64&+0.06&+0.65&+0.00&+0.33\\
\hline
\end{tabular}
}
\end{small}
\vspace{-0ex}
\end{table*}

\header
{\bf Centrality-based sampling} selects the $b$ edges according to their centrality values in $\G$. Recall in Theorem~\ref{lem:centrality} that the $L_2$ norm $\|\ZM^{(\textrm{c})}_{e_{u,i}}\|^2_2$ of each centrality embedding $\ZM^{(\textrm{c})}$ approximates the spanning centrality of $e_{u,i}$. As such, we sample the adjacent edges of $u$ and $i$ with a probability to $\textstyle \frac{\|\ZM^{(\textrm{c})}_{e_{u,x}}\|^2_2}{\sum_{\ell\in \N(u)}{\|\ZM^{(\textrm{c})}_{e_{u,\ell}}\|^2_2}}\ \text{and}\ \frac{\|\ZM^{(\textrm{c})}_{e_{v,i}}\|^2_2}{\sum_{\mu\in \N(i)}{\|\ZM^{(\textrm{c})}_{e_{\mu,i}}\|^2_2}}$,
respectively.

\section{Experiments}

This section experimentally evaluates \algo against 11 baselines in terms of TIC performance over 8 real datasets, followed by ablation studies. More details regarding TIN datasets, baselines, hyperparameters, and additional experimental results can be found in Appendix \ref{sec:additional_details}.
All the experiments are conducted on a Linux machine with an NVIDIA A100 GPU(80GB RAM), AMD EPYC 7513 CPU (2.6 GHz), and 1TB RAM. 
The source code is publicly accessible at \url{https://github.com/HKBU-LAGAS/SAFT}.

\begin{table*}[!t] 
\centering
\renewcommand{\arraystretch}{0.9}
\caption{TIC performance on medium/large datasets. The best results are bolded, while the best baselines are \underline{underlined}. OOT (out-of-time) indicates the method cannot report results within 1 day.}
\label{tab:large_datasets}
\vspace{-2ex}
\begin{small}
\addtolength{\tabcolsep}{-0.25em}
\resizebox{0.9\textwidth}{!}{
\begin{tabular}{ c | c  c  | c  c  | c  c  | c  c  }
\hline
\multirow{2}{*}{\textbf{Model}} & \multicolumn{2}{ c |}{\textbf{Goodreads-Poetry}} & \multicolumn{2}{ c |}{\textbf{Google-Vermont}} & \multicolumn{2}{ c |}{\textbf{Google-Hawaii}} & \multicolumn{2}{ c }{\textbf{Amazon-Products}}  \\ 
\cline{2-9}
& {Macro-F1} \textuparrow & {Micro-F1} \textuparrow & {Macro-F1} \textuparrow & {Micro-F1} \textuparrow& {Macro-F1} \textuparrow & {Micro-F1} \textuparrow& {Macro-F1} \textuparrow & {Micro-F1} \textuparrow\\
\hline

TF-IDF~\cite{robertson1994some}  & 33.31 & 46.72 & 46.66 & 68.68 & 44.14 & 66.37 & 45.57 & 76.03 \\ 
TF-IDF+NODES & 34.43 & 47.30 & 49.42 & 71.27 & 49.80 & 71.56 & 52.16 & 76.98 \\ \hline
AttrE2vec~\cite{bielak2022attre2vec}  & OOT & OOT & OOT & OOT & OOT & OOT & OOT & OOT \\
TER+AER (PPR)~\cite{wang2023efficient} & 23.34 & 43.56 & 35.93 & 65.11 & 32.33 & 63.31 & 25.29 & 71.86 \\
TER+AER (HKPR)~\cite{wang2023efficient} & 23.24 & 43.58 & 33.46 & 64.74 & 31.08 & 63.41 & 24.72 & 71.79 \\
EAGLE~\cite{wang2024effective} & 16.76 & 41.17 & 29.60 & 63.24 & 22.59 & 61.45 & 17.07 & 71.35 \\  \hline

BERT~\cite{devlin2018bert}  & 39.94 & 49.18 & 54.31 & 70.49 & 52.63 & 69.57 & 58.91 & 80.10 \\

BERT+NODES & 40.14 & 49.78 & 54.46 & 70.58 & 51.90 & 69.37 & 59.48 & 79.47 \\

BERT+GraphSAGE~\cite{hamilton2017inductive} & 40.53 & 51.36 & 55.31 & 70.32 & 53.46 & 69.55 & 60.00 & 80.56 \\
Deberta~\cite{he2020deberta} + GCN & 40.01 & 51.30 & 54.56 & 71.32 & 55.93 & 73.45 & 60.96 & 80.80 \\
Sentence-BERT~\cite{reimers2019sentence} + GCN & 40.10 & 51.32 & 55.08 & 71.40 & 56.08 & 73.50 & 61.10 & 81.00 \\
LLaMA-2 (7B)~\cite{touvron2023llama} + GCN & 39.21 & 50.20 & 49.73 & 66.90 & 54.31 & 72.80 & 60.35 & 80.00 \\
\hline
GraphFormers~\cite{yang2021graphformers} & 40.38 & 51.15 & 54.75 & 70.17 & 53.33 & 69.83 & 60.50 & 80.53 \\ 
GIANT~\cite{chien2021node} & 40.08 & 51.10 & 55.12 & 71.10 & 56.11 & 73.20 & 61.05 & 80.90 \\
GLEM-LLM~\cite{chen2024exploring} & 40.12 & 51.20 & 55.34 & 71.20 &56.32& 73.30 &61.11&81.00 \\
GLEM-GNN~\cite{chen2024exploring} & 40.29 & 51.25 & \underline{56.21} & 72.00 &56.78&73.60& \underline{61.28} &\underline{81.10} \\
Edgeformers~\cite{jin2023edgeformers} & \underline{40.64} & \underline{51.43} & 56.09 & \underline{72.49} & \underline{56.98} & \underline{74.68} & 61.22 & 81.07 \\ \hline
    
\algognn & \textbf{41.55} & \textbf{52.44} & \textbf{57.77} & \textbf{73.81} & 57.51 & \textbf{74.87} & 61.45 & 81.26 \\
Improv. & +0.91 & +1.01 & +1.56 & +1.32 & +0.53 & +0.19 & +0.17 & +0.16 \\
\algogau & 41.43 & 52.43 & 56.60 & 73.45 & \textbf{57.72} & 74.73 & \textbf{61.58} & \textbf{81.37} \\ 
Improv. & +0.79 & +1.00 & +0.39 & +0.96 & +0.74 & +0.05 & +0.30 & +0.27\\
\hline
\end{tabular}
}
\end{small}
\end{table*}

\begin{table*}[!t]
    \centering
    \renewcommand{\arraystretch}{0.9} %
    \caption{Ablation study on SAFT.}
    \label{tab:combined_ablation_study}
    \vspace{-3ex} %
    \begin{small}
    \addtolength{\tabcolsep}{-0.25em} %
    \resizebox{0.9\textwidth}{!}{
        \begin{tabular}{ c | c | c c | c c | c c | c c }
            \hline
            \multirow{2}{*}{\textbf{Method}} & \multirow{2}{*}{\textbf{Variant}} & \multicolumn{2}{c|}{\textbf{Goodreads-Children}} & \multicolumn{2}{c|}{\textbf{Amazon-Apps}} & \multicolumn{2}{c|}{\textbf{Goodreads-Poetry}} & \multicolumn{2}{c}{\textbf{Google-Vermont}} \\ \cline{3-10}
            & & Macro-F1 \textuparrow & Micro-F1 \textuparrow & Macro-F1 \textuparrow & Micro-F1 \textuparrow & Macro-F1 \textuparrow & Micro-F1 \textuparrow & Macro-F1 \textuparrow & Micro-F1 \textuparrow \\ 
            \hline
            \multirow{6}{*}{LGA} 
            & Full Model & \textbf{36.95} & \textbf{51.68} & \textbf{45.92} & \textbf{61.06} & \textbf{41.55} & \textbf{52.44} & \textbf{57.77} & \textbf{73.81} \\
            & w/o User-wise MP & 35.68 & 48.82 & 44.38 & 59.47 & 40.52 & 51.06 & 56.04 & 72.98  \\
            & w/o Item-wise MP & 36.13 & 49.26 & 44.11 & 57.40 & 40.70 & 51.48 & 56.27 & 72.99  \\
            & w/o User- and Item-wise MP & 35.00 & 50.07 & 45.48 & 58.68 & 39.80 & 49.44 & 56.62 & 72.08 \\
            & w/o Distance Embeddings & 36.43 & 49.82 & 43.87 & 57.15 & 40.82 & 52.17 & 55.79 & 71.93 \\
            & w/o Centrality Embeddings & 35.12 & 49.77 & 44.89 & 57.77 & 40.93 & 52.04 & 56.71 & 72.63 \\
            \hline
            \multirow{6}{*}{GAU} 
            & Full Model & \textbf{36.98} & \textbf{51.17} & \textbf{45.07} & \textbf{59.06} & \textbf{41.43} & \textbf{52.43} & \textbf{56.60} & \textbf{73.45}  \\
            & w/o User-wise MP & 35.55 & 50.10 & 44.31 & 56.95 & 40.20 & 50.66 & 56.02 & 72.35   \\
            & w/o Item-wise MP & 35.59 & 50.07 & 43.24 & 55.87 & 40.99 & 51.02 & 56.28 & 73.22   \\
            & w/o User- and Item-wise MP & 35.53 & 50.25 & 44.31 & 56.14 & 40.40 & 50.62 & 56.34 & 73.12   \\
            & w/o Distance Embeddings & 35.38 & 49.33 & 44.08 & 57.03 & 40.21 & 50.99 & 56.15 & 73.04   \\
            & w/o Centrality Embeddings & 35.38 & 49.33 & 44.08 & 57.03 & 40.82 & 51.56 & 56.17 & 72.80   \\
            \hline
        \end{tabular}
    }
    \end{small}
    \label{tab:combined_ablation_study}
\end{table*}

\subsection{Experiment Setting}
\stitle{\bf Datasets}
We experiment with 8 real-world TINs collected from Amazon~\cite{he2016ups}, Goodreads~\cite{wan2019fine}, and Google~\cite{li2022uctopic}, in which items correspond to Amazon products, Goodreads books, and Google local businesses, respectively, and textual interactions are reviews from users. Table \ref{tab:dataset_statistics} shows the statistics of these datasets.

\stitle{\bf Baselines}
We evaluate \algo{} against 17 competitors. They can be categorized into four groups: 
\begin{itemize}[leftmargin=*]
\item 2 bag-of-words methods, TF-IDF~\cite{robertson1994some} and TF-IDF+NODES;
\item 4 edge-wise representation learning methods: AttrE2Vec~\cite{bielak2022attre2vec}, EAGLE~\cite{wang2024effective}, and two versions of TER+AER~\cite{wang2023efficient};
\item BERT~\cite{devlin2018bert} and 5 cascade models that combine PLMs and node-wise GNNs, including BERT+NODES~\cite{jin2023edgeformers}, BERT+GraphSAGE~\cite{hamilton2017inductive}, Deberta~\cite{he2020deberta}+GCN, Sentence-BERT~\cite{reimers2019sentence}+GCN, and LLaMA-2 (7B)~\cite{touvron2023llama}+GCN;
\item 5 sophisticated models combining PLMs and node-wise GNNs, i.e., Graphformers~\cite{yang2021graphformers}, Edgeformers~\cite{jin2023edgeformers}, GLEM-LLM, GLEM-GNN~\cite{zhao2022learning, chen2024exploring}, and GIANT~\cite{chien2021node, chen2024exploring}.
\end{itemize}
Specifically, as in ~\cite{jin2023edgeformers}, TF-IDF+NODES and BERT+NODES incorporate node embeddings to inject TIN structural information.

\subsection{Classification Performance}
Tables \ref{tab:small_datasets} and \ref{tab:large_datasets} present the TIC performance of \algo and baselines across TINs. We observe that:
(1) Notably, \algo consistently outperforms baselines across all TIN sizes, with \algognn achieving the highest Micro-F1 (61.06) on Amazon-Apps and \algogau reaching the best Macro-F1 (57.72) on Google-Hawaii, showcasing their ability to combine semantic embeddings with structural information effectively.
(2) On small TINs (Table \ref{tab:small_datasets}), experiments with BERT-Tiny and full-batch processing show that static methods like TF-IDF and AttrE2vec underperform due to their inability to capture structural patterns. In contrast, models focusing on TIN structure, such as Graphformers and \algo, perform better. For example, Graphformers achieves Macro-F1 (36.51) on Goodreads-Crime, highlighting the importance of structural features.
(3) On medium and large TINs (Table \ref{tab:large_datasets}), BERT-Base with mini-batch processing and graph sampling demonstrate the benefits of integrating rich textual data. Transformer-based models, such as Edgeformers and \algo, show strong performance by leveraging contextualized semantics. For instance, on Goodreads-Poetry, \algo achieves a Micro-F1 of 52.44, outperforming Edgeformers with an improvement of +1.01.

\vspace{-2ex}
\subsection{Ablation Study}
\stitle{Key Components in SAFT}
We perform ablation studies to evaluate the impact of key components in SAFT, including User-wise and Item-wise Message Passing (MP), Distance Embeddings, and Centrality Embeddings. We make observations from Tables~\ref{tab:combined_ablation_study}:
(1) User-wise and Item-wise MP: This mechanism is crucial for capturing interaction features. Removing both User-wise and Item-wise MP leads to significant drops in performance. For instance, in \algognn~on Goodreads-Children, Macro-F1 decreases from 36.95\% to 35.00\% (-1.95\%).
(2): Distance Embeddings: This part captures the local structural context. On Amazon-Apps, removing Distance Embeddings in \algognn~reduces Macro-F1 from 45.92\% to 43.87\% (-2.05\%), demonstrating the importance of distance information.
(3)Centrality Embeddings: This component involves interaction importance. In \algognn~on Google-Vermont, removing Centrality Embeddings decreases Macro-F1 from 57.77\% to 56.71\% (-1.06\%), highlighting its role in improving performance.

\stitle{Other Distance and Centrality Embeddings}
We further evaluate SAFT with distance constructed from node2vec \cite{grover2016node2vec}, shortest path distance (SPD) \cite{madkour2017survey}, and common neighbors (CN) \cite{yang2016predicting}, and centrality embeddings constructed based on personalized PageRank (PPR)~\cite{yang2022efficient}, commute time \cite{qiu2007clustering}, and SimRank \cite{jeh2002simrank}. As displayed in Table \ref{tab:comparison_distance_embeddings} and Table \ref{tab:comparison_centrality_embeddings}, it can be observed that our resistance distance-based embeddings and spanning centrality-based embeddings consistently achieve higher performance in TIC, indicating the effectiveness of our structural encoding techniques in Section \ref{sec:struct-emb}. The superiority of our structural encoding lies in its design for edges in TINs, which are essentially bipartite, capturing the unique topological characteristics of such edges. Our theoretical analyses in Theorems \ref{lem:Katz} and \ref{lem:centrality} further reveal their capabilities in preserving the structural properties of the original graphs. In contrast, existing distance and centrality measures are primarily designed for nodes, overlooking the unique traits of edges in TINs.
\begin{table}[!h] 
\renewcommand{\arraystretch}{0.7}
\caption{\algo{} with various distance embeddings.}
\label{tab:comparison_distance_embeddings}
\vspace{-3ex}
\begin{center}
\small
\begin{tabular}{ l|l|c|c } 
\hline
{\bf Dataset} & {\bf Method} & {\bf Macro (F1)} & {\bf Micro (F1)} \\ 
\hline
\multirow{4}{*}{Amazon-Apps}           & Ours                  & {\bf 45.92} & {\bf 61.06} \\ 
                      & node2vec              & 44.85 & 60.12 \\ 
                      & SPD & 44.55 & 59.98 \\ 
                      & CN        & 43.78 & 59.20 \\ 
\hline
\multirow{4}{*}{Google-Vermont}        & Ours                  & {\bf 57.77} & {\bf 73.81} \\ 
                      & node2vec              & 56.98 & 72.95 \\ 
                      & SPD & 56.40 & 72.45 \\ 
                      & CN        & 56.00 & 72.15 \\ 
\hline
\end{tabular}
\end{center}
\vspace{-3ex}
\end{table}

\begin{table}[!h] 
\renewcommand{\arraystretch}{0.9}
\caption{\algo{} with various centrality embeddings.}
\label{tab:comparison_centrality_embeddings}
\vspace{-3ex}
\begin{center}
\small
\resizebox{\columnwidth}{!}{
\begin{tabular}{ l|l|c|c } 
\hline
{\bf Dataset} & {\bf Method} & {\bf Macro (F1)} & {\bf Micro (F1)} \\ 
\hline
\multirow{4}{*}{Amazon-Apps}           & Ours                  & {\bf 45.92} & {\bf 61.06} \\ 
                      & PPR                   & 45.10 & 60.50 \\ 
                      & Commute Time          & 44.40 & 59.85 \\ 
                      & SimRank               & 44.85 & 60.20 \\ 
\hline
\multirow{4}{*}{Google-Vermont}        & Ours                  & {\bf 57.77} & {\bf 73.81} \\ 
                      & PPR                   & 57.00 & 73.20 \\ 
                      & Commute Time          & 56.50 & 72.60 \\ 
                      & SimRank               & 56.75 & 72.90 \\ 
\hline
\end{tabular}
}
\end{center}
\vspace{-3ex}
\end{table}

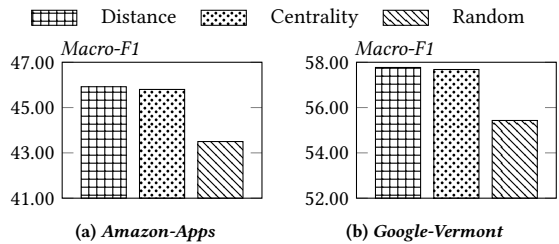
\begin{figure}[!t]
\centering
\begin{small}
\begin{tikzpicture}
\begin{customlegend}[
        legend entries={{Distance},{Centrality}, {Random}},
        legend columns=4,
        area legend,
        legend style={at={(0.45,1.15)},anchor=north,draw=none,font=\small,column sep=0.25cm}]
        \addlegendimage{ pattern={grid}}  
        \addlegendimage{ pattern={crosshatch dots}}   
        \addlegendimage{pattern=north west lines} 
    \end{customlegend}
\end{tikzpicture}
\\[-\lineskip]
\vspace{-4mm}
\subfloat[{\em Amazon-Apps}]{
\begin{tikzpicture}[scale=1]
\begin{axis}[
    height=\columnwidth/2.5,
    width=\columnwidth/2.0,
    xtick=\empty,
    ybar=5.0pt,
    bar width=0.6cm,
    enlarge x limits=true,
    ylabel={\em Macro-F1},
    xticklabel=\empty,
    ymin=41.00,
    ymax=47.00,
    ytick={41.00,43.00,45.00,47.00},
    yticklabels={41.00,43.00,45.00,47.00},
    xticklabel style = {font=\small},
    yticklabel style = {font=\small},
    every axis y label/.style={at={(current axis.north west)},right=5mm,above=0mm},
    legend style={draw=none, at={(1.02,1.02)},anchor=north west,cells={anchor=west},font=\small},
    legend image code/.code={ \draw [#1] (0cm,-0.1cm) rectangle (0.3cm,0.15cm); },
    ]

\addplot [ pattern={grid}] coordinates {(1,45.92) }; 
\addplot [ pattern={crosshatch dots}] coordinates {(1,45.80) }; 
\addplot [pattern=north west lines] coordinates {(1,43.50) }; 

\end{axis}
\end{tikzpicture}\hspace{4mm}\label{fig:time-photos}%
}%
\subfloat[{\em Google-Vermont}]{
\begin{tikzpicture}[scale=1]
\begin{axis}[
    height=\columnwidth/2.5,
    width=\columnwidth/2.0,
    xtick=\empty,
    ybar=5.0pt,
    bar width=0.6cm,
    enlarge x limits=true,
    ylabel={\em Macro-F1},
    xticklabel=\empty,
    ymin=52.00,
    ymax=58.00,
    ytick={52.00,54.00,56.00,58.00},
    yticklabels={52.00,54.00,56.00,58.00},
    xticklabel style = {font=\small},
    yticklabel style = {font=\small},
    every axis y label/.style={at={(current axis.north west)},right=5mm,above=0mm},
    legend style={draw=none, at={(1.02,1.02)},anchor=north west,cells={anchor=west},font=\small},
    legend image code/.code={ \draw [#1] (0cm,-0.1cm) rectangle (0.3cm,0.15cm); },
    ]

\addplot [ pattern={grid}] coordinates {(1,57.77) }; 
\addplot [ pattern={crosshatch dots}] coordinates {(1,57.68) }; 
\addplot [pattern=north west lines] coordinates {(1,55.43) }; 

\end{axis}
\end{tikzpicture}\hspace{0mm}\label{fig:time-Cora}%
}%
\end{small}
\vspace{-2ex}
\caption{SAFT with various graph sampling strategies} \label{fig:comparison_sampling}
\vspace{-3ex}
\end{figure}

\stitle{Sampling Strategy}
To show the effectiveness of our sampling strategies for mini-batch training, we compare our distance and centrality-based sampling techniques depicted in Section \ref{sec:graphSampling} with random sampling methods. Fig. \ref{fig:comparison_sampling} highlights that distance and centrality-based sampling outperform random sampling by a significant margin, demonstrating their importance in enhancing model performance and scalability.

\stitle{Other Backbone PLMs}
We evaluate SAFT with different language models and compare the performance with the best competitors. The results, presented in Table \ref{tab:comparison_PLMs}, indicate that SAFT consistently outperforms Graphformers and Edgeformers when equipped with the same PLM (e.g., Roberta \cite{liu2019roberta}) and validate that its improvements are model-agnostic and not dependent on a specific PLM, confirming its flexibility and effectiveness in leveraging advanced PLMs. 

\begin{table}[!h] 
\renewcommand{\arraystretch}{0.9}
\caption{Evaluation with other PLM backbone (i.e., Roberta).}
\label{tab:comparison_PLMs}
\vspace{-3ex}
\begin{center}
\small
\begin{tabular}{ l|l|c|c } 
\hline
{\bf Dataset} & {\bf Method} & {\bf Macro (F1)} & {\bf Micro (F1)} \\ 
\hline
\multirow{3}{*}{Amazon-Apps}   & \algo{}        & {\bf 47.30} & {\bf 63.50} \\ 
              & Edgeformers & 46.40 & 62.70 \\ 
              & Graphformers & 45.95 & 62.20 \\ 
\hline
\multirow{3}{*}{Google-Vermont}       & \algo{}        & {\bf 59.68} & {\bf 75.21} \\ 
              & Edgeformers & 58.49 & 74.07 \\ 
              & Graphformers & 57.85 & 73.50 \\ 
\hline
\end{tabular}
\end{center}
\vspace{-2ex}
\end{table}

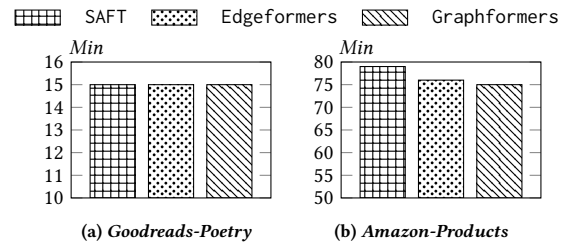
\begin{figure}[!t]
\centering
\begin{small}
\begin{tikzpicture}
\begin{customlegend}[
        legend entries={\texttt{SAFT},\texttt{Edgeformers}, \texttt{Graphformers}},
        legend columns=4,
        area legend,
        legend style={at={(0.45,1.15)},anchor=north,draw=none,font=\small,column sep=0.25cm}]
        \addlegendimage{ pattern={grid}}  
        \addlegendimage{ pattern={crosshatch dots}}   
        \addlegendimage{pattern=north west lines} 
    \end{customlegend}
\end{tikzpicture}
\\[-\lineskip]
\vspace{-4mm}
\subfloat[{\em Goodreads-Poetry}]{
\begin{tikzpicture}[scale=1]
\begin{axis}[
    height=\columnwidth/2.5,
    width=\columnwidth/2.0,
    xtick=\empty,
    ybar=5.0pt,
    bar width=0.6cm,
    enlarge x limits=true,
    ylabel={\em Min},
    xticklabel=\empty,
    ymin=10,
    ymax=16,
    ytick={10,11,12,13,14,15,16},
    yticklabels={10,11,12,13,14,15,16},
    xticklabel style = {font=\small},
    yticklabel style = {font=\small},
    every axis y label/.style={at={(current axis.north west)},right=2mm,above=0mm},
    legend style={draw=none, at={(1.02,1.02)},anchor=north west,cells={anchor=west},font=\small},
    legend image code/.code={ \draw [#1] (0cm,-0.1cm) rectangle (0.3cm,0.15cm); },
    ]

\addplot [ pattern={grid}] coordinates {(1,15) }; 
\addplot [ pattern={crosshatch dots}] coordinates {(1,15) }; 
\addplot [pattern=north west lines] coordinates {(1,15) }; 

\end{axis}
\end{tikzpicture}\hspace{4mm}\label{fig:time-photos}%
}%
\subfloat[{\em Amazon-Products}]{
\begin{tikzpicture}[scale=1]
\begin{axis}[
    height=\columnwidth/2.5,
    width=\columnwidth/2.0,
    xtick=\empty,
    ybar=5.0pt,
    bar width=0.6cm,
    enlarge x limits=true,
    ylabel={\em Min},
    xticklabel=\empty,
    ymin=50,
    ymax=80,
    ytick={50,55,60,65,70,75,80},
    yticklabels={50,55,60,65,70,75,80},
    xticklabel style = {font=\small},
    yticklabel style = {font=\small},
    every axis y label/.style={at={(current axis.north west)},right=2mm,above=0mm},
    legend style={draw=none, at={(1.02,1.02)},anchor=north west,cells={anchor=west},font=\small},
    legend image code/.code={ \draw [#1] (0cm,-0.1cm) rectangle (0.3cm,0.15cm); },
    ]

\addplot [pattern={grid}] coordinates {(1,79) }; 
\addplot [pattern={crosshatch dots}] coordinates {(1,76) }; 
\addplot [pattern=north west lines] coordinates {(1,75) }; 

\end{axis}
\end{tikzpicture}\hspace{0mm}\label{fig:time-Cora}%
}%
\end{small}
\vspace{-2ex}
\caption{Training time of SAFT and baselines.} \label{fig:comparison_time}
\vspace{-3ex}
\end{figure}

\stitle{Training Time}
We compare the empirical training time per epoch of SAFT with competitors Edgeformers and Graphformers across two datasets. The results, shown in Fig. \ref{fig:comparison_time}, confirm that SAFT maintains competitive efficiency without sacrificing expressiveness or performance and that its training efficiency is comparable to these state-of-the-art methods, running in time linear to the size of the input graph.

\section{Conclusion}
In this paper, we proposed \algo, a novel architecture that integrates language- and graph-based models to address limitations in TIC. By jointly leveraging PLMs and LGA/GAUs, \algo effectively fuses textual semantics and structural features inherent to TINs. The model incorporates a proxy token to bridge microscopic (token-level) and macroscopic (interaction-level) signals, while structural embeddings based on centrality and resistance distance capture global and local topological patterns. Additionally, efficient graph sampling strategies selectively aggregate interaction information. Extensive evaluations on real-world TINs demonstrate that \algo consistently outperforms state-of-the-art methods in TIC accuracy. Future work will focus on refining structural embeddings and exploring broader graph-based applications.

\begin{acks}
This work is supported by the National Natural Science Foundation of China (No. 62302414), the Hong Kong RGC ECS grant (No. 22202623), Hong Kong RGC R1015-23, the Guangdong Basic and Applied Basic Research Foundation (Project No. 2023B1515130002), and the Huawei Gift Fund.
\end{acks}

\balance
\pagebreak
\bibliographystyle{ACM-Reference-Format}
\bibliography{sample-base}
\appendix
\section{Additional Experimental Details}
\label{sec:additional_details}
\begin{table}[H]
    \centering
    \caption{URLs of baseline codes.}
    \resizebox{0.45\textwidth}{!}{%
    \fontsize{5}{8}\selectfont
    \setlength{\arrayrulewidth}{0.01mm} %
    \begin{tabular}{|l|l|}
        \hline
        \textbf{Methods} & \textbf{URL} \\ \hline
        AttrE2vec & \url{https://github.com/attre2vec/attre2vec} \\ 
        GraphSAGE & \url{https://github.com/williamleif/GraphSAGE} \\ 
        TER+AER & \url{https://github.com/wanghewen/TER_AER} \\ 
        EAGLE & \url{https://github.com/wanghewen/EAGLE} \\ 
        GraphFormers & \url{https://github.com/microsoft/GraphFormers} \\ 
        Edgeformers & \url{https://github.com/PeterGriffinJin/Edgeformers} \\ \hline
    \end{tabular}
    \label{tab:baseline_codes}
    }
\end{table}

\begin{figure*}[!t]
\centering
\begin{small}
\begin{tikzpicture}
    \begin{customlegend}
    [legend columns=2,legend entries={Macro-F1,Micro-F1},
        legend style={at={(0.5,-0.1)},anchor=north,draw=none,font=\footnotesize,column sep=0.2cm}]
    \addlegendimage{line width=0.3mm, mark=o,color=orange}
    \addlegendimage{line width=0.3mm, mark=square,color=blue}
    \end{customlegend}
\end{tikzpicture}
\subfloat[Goodreads-Children]{
    \begin{tikzpicture}[scale=1,every mark/.append style={mark size=2pt}]
        \begin{axis}[
            height=\columnwidth/2.5,
            width=\columnwidth/2.1,
            ylabel={\it Macro-F1},
            xmin=0.5, xmax=5.5,
            ymin=35.8, ymax=37,
            xtick={1,2,3,4,5},
            ytick={35.8, 36.1, 36.4, 36.7, 37.0},
            yticklabel style = {font=\scriptsize},
            xticklabel style = {font=\scriptsize},
            xticklabels={1, 2, 3, 4, 5},
            yticklabels={35.8, 36.1, 36.4, 36.7, 37.0},
            every axis y label/.style={font=\footnotesize,at={(current axis.north west)},right=2mm,above=0mm},
            legend style={fill=none,font=\small,at={(0.02,0.99)},anchor=north west,draw=none},
        ]
        \addplot[line width=0.3mm, mark=o,color=orange]
            plot coordinates {
                (1, 35.84)
                (2, 35.92)
                (3, 36.95)
                (4, 35.88)
                (5, 35.89)
            };
        \end{axis}

        \begin{axis}[
            height=\columnwidth/2.5,
            width=\columnwidth/2.1,
            ylabel={\it Micro-F1},
            xmin=0.5, xmax=5.5,
            ymin=50.0, ymax=52.0,
            axis y line*=right,
            ylabel near ticks,
            ytick={50.0, 50.5, 51.0, 51.5, 52.0},
            yticklabel style = {font=\scriptsize},
            xmajorticks=false,
            yticklabels={50.0, 50.5, 51.0, 51.5, 52.0},
            enlargelimits=false,
            every axis y label/.style={font=\footnotesize,at={(current axis.north west)},right=17mm,above=0mm},
            legend style={fill=none,font=\small,at={(0.02,0.99)},anchor=north west,draw=none},
        ]
        \addplot[line width=0.3mm, mark=square,color=blue]
            plot coordinates {
                (1, 50.33)
                (2, 50.34)
                (3, 51.68)
                (4, 50.28)
                (5, 50.29)
            };
        \end{axis}
    \end{tikzpicture}
    \hspace{2mm}
}
\subfloat[Amazon-Apps]{
    \begin{tikzpicture}[scale=1,every mark/.append style={mark size=2pt}]
        \begin{axis}[
            height=\columnwidth/2.5,
            width=\columnwidth/2.1,
            ylabel={\it Macro-F1},
            xmin=0.5, xmax=5.5,
            ymin=44.4, ymax=46,
            xtick={1,2,3,4,5},
            ytick={44.4, 44.8, 45.2, 45.6, 46},
            yticklabel style = {font=\scriptsize},
            xticklabel style = {font=\scriptsize},
            xticklabels={1, 2, 3, 4, 5},
            yticklabels={44.4, 44.8, 45.2, 45.6, 46},
            every axis y label/.style={font=\footnotesize,at={(current axis.north west)},right=2mm,above=0mm},
            legend style={fill=none,font=\small,at={(0.02,0.99)},anchor=north west,draw=none},
        ]
        \addplot[line width=0.3mm, mark=o,color=orange]
            plot coordinates {
                (1, 45.92)
                (2, 45.91)
                (3, 45.91)
                (4, 44.77)
                (5, 44.77)
            };
        \end{axis}

        \begin{axis}[
            height=\columnwidth/2.5,
            width=\columnwidth/2.1,
            ylabel={\it Micro-F1},
            xmin=0.5, xmax=5.5,
            ymin=57.5, ymax=61.5,
            axis y line*=right,
            ylabel near ticks,
            ytick={57.5, 58.5, 59.5, 60.5, 61.5},
            yticklabel style = {font=\scriptsize},
            xmajorticks=false,
            yticklabels={57.5, 58.5, 59.5, 60.5, 61.5},
            enlargelimits=false,
            every axis y label/.style={font=\footnotesize,at={(current axis.north west)},right=17mm,above=0mm},
            legend style={fill=none,font=\small,at={(0.02,0.99)},anchor=north west,draw=none},
        ]
        \addplot[line width=0.3mm, mark=square,color=blue]
            plot coordinates {
                (1, 61.06)
                (2, 61.05)
                (3, 61.05)
                (4, 57.74)
                (5, 57.75)
            };
        \end{axis}
    \end{tikzpicture}
    \hspace{2mm}
}
\subfloat[Goodreads-Poetry]{
    \begin{tikzpicture}[scale=1,every mark/.append style={mark size=2pt}]
        \begin{axis}[
            height=\columnwidth/2.5,
            width=\columnwidth/2.1,
            ylabel={\it Macro-F1},
            xmin=0.5, xmax=5.5,
            ymin=40.5, ymax=41.7,
            xtick={1,2,3,4,5},
            ytick={40.5, 40.8, 41.1, 41.4, 41.7},
            yticklabel style = {font=\scriptsize},
            xticklabel style = {font=\scriptsize},
            xticklabels={1, 2, 3, 4, 5},
            yticklabels={40.5, 40.8, 41.1, 41.4, 41.7},
            every axis y label/.style={font=\footnotesize,at={(current axis.north west)},right=2mm,above=0mm},
            legend style={fill=none,font=\small,at={(0.02,0.99)},anchor=north west,draw=none},
        ]
        \addplot[line width=0.3mm, mark=o,color=orange]
            plot coordinates {
                (1, 41.01)
                (2, 40.85)
                (3, 40.85)
                (4, 41.55)
                (5, 40.68)
            };
        \end{axis}

        \begin{axis}[
            height=\columnwidth/2.5,
            width=\columnwidth/2.1,
            ylabel={\it Micro-F1},
            xmin=0.5, xmax=5.5,
            ymin=51.8, ymax=52.5,
            axis y line*=right,
            ylabel near ticks,
            ytick={51.8, 52, 52.2, 52.4, 52.5},
            yticklabel style = {font=\scriptsize},
            xmajorticks=false,
            yticklabels={51.8, 52, 52.2, 52.4, 52.5},
            enlargelimits=false,
            every axis y label/.style={font=\footnotesize,at={(current axis.north west)},right=17mm,above=0mm},
            legend style={fill=none,font=\small,at={(0.02,0.99)},anchor=north west,draw=none},
        ]
        \addplot[line width=0.3mm, mark=square,color=blue]
            plot coordinates {
                (1, 52.31)
                (2, 51.98)
                (3, 52.29)
                (4, 52.44)
                (5, 52.12)
            };
        \end{axis}
    \end{tikzpicture}
    \hspace{2mm}
}
\subfloat[Google-Vermont]{
    \begin{tikzpicture}[scale=1,every mark/.append style={mark size=2pt}]
        \begin{axis}[
            height=\columnwidth/2.5,
            width=\columnwidth/2.1,
            ylabel={\it Macro-F1},
            xmin=0.5, xmax=5.5,
            ymin=56.4, ymax=58.2,
            xtick={1,2,3,4,5},
            ytick={56.4, 57.0, 57.6, 58.2},
            yticklabel style = {font=\scriptsize},
            xticklabel style = {font=\scriptsize},
            xticklabels={1, 2, 3, 4, 5},
            yticklabels={56.4, 57.0, 57.6, 58.2},
            every axis y label/.style={font=\footnotesize,at={(current axis.north west)},right=2mm,above=0mm},
            legend style={fill=none,font=\small,at={(0.02,0.99)},anchor=north west,draw=none},
        ]
        \addplot[line width=0.3mm, mark=o,color=orange]
            plot coordinates {
                (1, 56.63)
                (2, 57.77)
                (3, 56.82)
                (4, 56.93)
                (5, 56.42)
            };
        \end{axis}

        \begin{axis}[
            height=\columnwidth/2.5,
            width=\columnwidth/2.1,
            ylabel={\it Micro-F1},
            xmin=0.5, xmax=5.5,
            ymin=72.4, ymax=73.9,
            axis y line*=right,
            ylabel near ticks,
            ytick={72.4, 72.9, 73.4, 73.9},
            yticklabel style = {font=\scriptsize},
            xmajorticks=false,
            yticklabels={72.4, 72.9, 73.4, 73.9},
            enlargelimits=false,
            every axis y label/.style={font=\footnotesize,at={(current axis.north west)},right=17mm,above=0mm},
            legend style={fill=none,font=\small,at={(0.02,0.99)},anchor=north west,draw=none},
        ]
        \addplot[line width=0.3mm, mark=square,color=blue]
            plot coordinates {
                (1, 72.68)
                (2, 73.81)
                (3, 73.31)
                (4, 73.00)
                (5, 73.24)
            };
        \end{axis}
    \end{tikzpicture}
    \hspace{2mm}
}
\end{small}
\vspace{-2ex}
\caption{Macro-F1 and Micro-F1 when varying \#layers} \label{fig:vary-layer}
\vspace{-2ex}
\end{figure*}
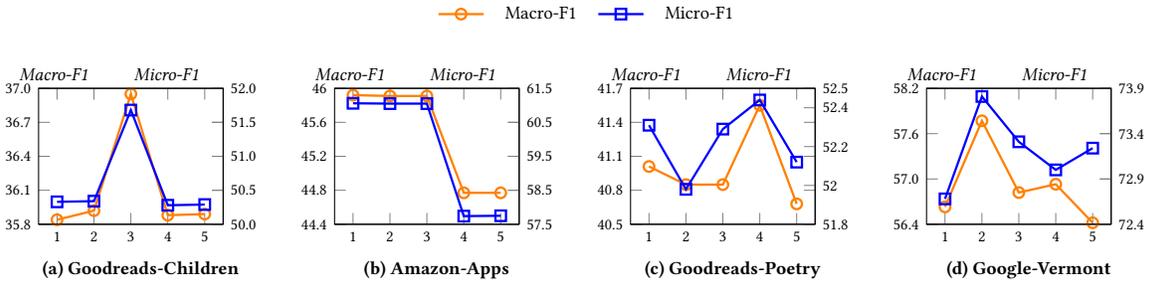
\begin{figure*}[!t]
\centering
\begin{small}
\subfloat[Goodreads-Children]{
    \begin{tikzpicture}[scale=1,every mark/.append style={mark size=2pt}]
        \begin{axis}[
            height=\columnwidth/2.5,
            width=\columnwidth/2.1,
            ylabel={\it Macro-F1},
            xmin=0.5, xmax=5.5,
            ymin=35, ymax=37,
            xtick={1,2,3,4,5},
            ytick={35, 36, 37},
            yticklabel style = {font=\scriptsize},
            xticklabel style = {font=\scriptsize},
            xticklabels={16,32,64,128,256},
            yticklabels={35, 36, 37},
            every axis y label/.style={font=\footnotesize,at={(current axis.north west)},right=2mm,above=0mm},
            legend style={fill=none,font=\small,at={(0.02,0.99)},anchor=north west,draw=none},
        ]
        \addplot[line width=0.3mm, mark=o,color=orange]
            plot coordinates {
                (1, 36.25)
                (2, 35.15)
                (3, 36.91)
                (4, 35.71)
                (5, 36.95)
            };
        \end{axis}

        \begin{axis}[
            height=\columnwidth/2.5,
            width=\columnwidth/2.1,
            ylabel={\it Micro-F1},
            xmin=0.5, xmax=5.5,
            ymin=49.75, ymax=51.75,
            axis y line*=right,
            ylabel near ticks,
            ytick={49.75, 50.75, 51.75},
            yticklabel style = {font=\scriptsize},
            xmajorticks=false,
            yticklabels={49.75, 50.75, 51.75},
            enlargelimits=false,
            every axis y label/.style={font=\footnotesize,at={(current axis.north west)},right=17mm,above=0mm},
            legend style={fill=none,font=\small,at={(0.02,0.99)},anchor=north west,draw=none},
        ]
        \addplot[line width=0.3mm, mark=square,color=blue]
            plot coordinates {
                (1, 49.96)
                (2, 50.17)
                (3, 50.57)
                (4, 49.94)
                (5, 51.68)
            };
        \end{axis}
    \end{tikzpicture}
    \hspace{2mm}
}
\subfloat[Amazon-Apps]{
    \begin{tikzpicture}[scale=1,every mark/.append style={mark size=2pt}]
        \begin{axis}[
            height=\columnwidth/2.5,
            width=\columnwidth/2.1,
            ylabel={\it Macro-F1},
            xmin=0.5, xmax=5.5,
            ymin=43, ymax=46,
            xtick={1,2,3,4,5},
            ytick={43, 44, 45, 46},
            yticklabel style = {font=\scriptsize},
            xticklabel style = {font=\scriptsize},
            xticklabels={16,32,64,128,256},
            yticklabels={43, 44, 45, 46},
            every axis y label/.style={font=\footnotesize,at={(current axis.north west)},right=2mm,above=0mm},
            legend style={fill=none,font=\small,at={(0.02,0.99)},anchor=north west,draw=none},
        ]
        \addplot[line width=0.3mm, mark=o,color=orange]
            plot coordinates {
                (1, 44.41)
                (2, 44.20)
                (3, 45.92)
                (4, 44.58)
                (5, 43.47)
            };
        \end{axis}

        \begin{axis}[
            height=\columnwidth/2.5,
            width=\columnwidth/2.1,
            ylabel={\it Micro-F1},
            xmin=0.5, xmax=5.5,
            ymin=56, ymax=62,
            axis y line*=right,
            ylabel near ticks,
            ytick={56, 58, 60, 62},
            yticklabel style = {font=\scriptsize},
            xmajorticks=false,
            yticklabels={56, 58, 60, 62},
            enlargelimits=false,
            every axis y label/.style={font=\footnotesize,at={(current axis.north west)},right=17mm,above=0mm},
            legend style={fill=none,font=\small,at={(0.02,0.99)},anchor=north west,draw=none},
        ]
        \addplot[line width=0.3mm, mark=square,color=blue]
            plot coordinates {
                (1, 57.45)
                (2, 58.02)
                (3, 61.06)
                (4, 58.30)
                (5, 56.38)
            };
        \end{axis}
    \end{tikzpicture}
    \hspace{2mm}
}
\subfloat[Goodreads-Poetry]{
    \begin{tikzpicture}[scale=1,every mark/.append style={mark size=2pt}]
        \begin{axis}[
            height=\columnwidth/2.5,
            width=\columnwidth/2.1,
            ylabel={\it Macro-F1},
            xmin=0.5, xmax=5.5,
            ymin=40.5, ymax=42,
            xtick={1,2,3,4,5},
            ytick={40.5, 41, 41.5, 42},
            yticklabel style = {font=\scriptsize},
            xticklabel style = {font=\scriptsize},
            xticklabels={16,32,64,128,256},
            yticklabels={40.5, 41, 41.5, 42},
            every axis y label/.style={font=\footnotesize,at={(current axis.north west)},right=2mm,above=0mm},
            legend style={fill=none,font=\small,at={(0.02,0.99)},anchor=north west,draw=none},
        ]
        \addplot[line width=0.3mm, mark=o,color=orange]
            plot coordinates {
                (1, 41.44)
                (2, 41.01)
                (3, 41.55)
                (4, 41.70)
                (5, 41.24)
            };
        \end{axis}

        \begin{axis}[
            height=\columnwidth/2.5,
            width=\columnwidth/2.1,
            ylabel={\it Micro-F1},
            xmin=0.5, xmax=5.5,
            ymin=51.5, ymax=53,
            axis y line*=right,
            ylabel near ticks,
            ytick={51.5, 52, 52.5, 53},
            yticklabel style = {font=\scriptsize},
            xmajorticks=false,
            yticklabels={51.5, 52, 52.5, 53},
            enlargelimits=false,
            every axis y label/.style={font=\footnotesize,at={(current axis.north west)},right=17mm,above=0mm},
            legend style={fill=none,font=\small,at={(0.02,0.99)},anchor=north west,draw=none},
        ]
        \addplot[line width=0.3mm, mark=square,color=blue]
            plot coordinates {
                (1, 51.99)
                (2, 52.31)
                (3, 52.44)
                (4, 52.59)
                (5, 52.39)
            };
        \end{axis}
    \end{tikzpicture}
    \hspace{2mm}
}
\subfloat[Google-Vermont]{
    \begin{tikzpicture}[scale=1,every mark/.append style={mark size=2pt}]
        \begin{axis}[
            height=\columnwidth/2.5,
            width=\columnwidth/2.1,
            ylabel={\it Macro-F1},
            xmin=0.5, xmax=5.5,
            ymin=56.0, ymax=57.8,
            xtick={1,2,3,4,5},
            ytick={56.0, 56.6, 57.2, 57.8},
            yticklabel style = {font=\scriptsize},
            xticklabel style = {font=\scriptsize},
            xticklabels={16,32,64,128,256},
            yticklabels={56.0, 56.6, 57.2, 57.8},
            every axis y label/.style={font=\footnotesize,at={(current axis.north west)},right=2mm,above=0mm},
            legend style={fill=none,font=\small,at={(0.02,0.99)},anchor=north west,draw=none},
        ]
        \addplot[line width=0.3mm, mark=o,color=orange]
            plot coordinates {
                (1, 56.32)
                (2, 57.04)
                (3, 57.77)
                (4, 56.95)
                (5, 56.82)
            };
        \end{axis}

        \begin{axis}[
            height=\columnwidth/2.5,
            width=\columnwidth/2.1,
            ylabel={\it Micro-F1},
            xmin=0.5, xmax=5.5,
            ymin=72.5, ymax=74,
            axis y line*=right,
            ylabel near ticks,
            ytick={72.5, 73, 73.5, 74},
            yticklabel style = {font=\scriptsize},
            xmajorticks=false,
            yticklabels={72.5, 73, 73.5, 74},
            enlargelimits=false,
            every axis y label/.style={font=\footnotesize,at={(current axis.north west)},right=17mm,above=0mm},
            legend style={fill=none,font=\small,at={(0.02,0.99)},anchor=north west,draw=none},
        ]
        \addplot[line width=0.3mm, mark=square,color=blue]
            plot coordinates {
                (1, 72.89)
                (2, 73.13)
                (3, 73.81)
                (4, 73.11)
                (5, 73.51)
            };
        \end{axis}
    \end{tikzpicture}
    \hspace{2mm}
}
\end{small}
\vspace{-2ex}
\caption{Macro-F1 and Micro-F1 when varying \#dim} \label{fig:vary-dim}
\vspace{-2ex}
\end{figure*}
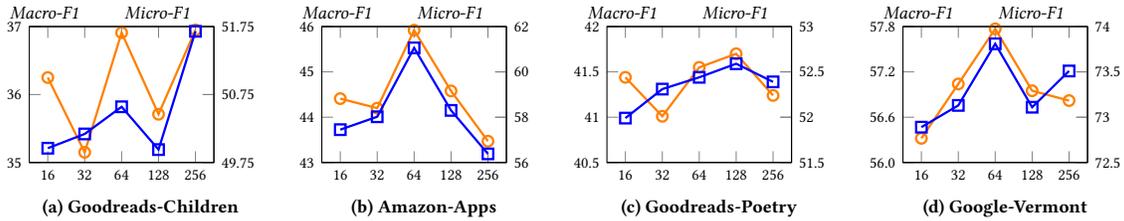
\begin{figure*}[!t]
\centering
\begin{small}
\subfloat[Goodreads-Children]{
    \begin{tikzpicture}[scale=1,every mark/.append style={mark size=2pt}]
        \begin{axis}[
            height=\columnwidth/2.5,
            width=\columnwidth/2.1,
            ylabel={\it Macro-F1},
            xmin=0.5, xmax=5.5,
            ymin=35.8, ymax=37.0,
            xtick={1,2,3,4,5},
            ytick={35.8, 36.1, 36.4, 36.7, 37.0},
            yticklabel style = {font=\scriptsize},
            xticklabel style = {font=\scriptsize},
            xticklabels={0.0, 0.5, 1.0, 1.5, 2.0},
            yticklabels={35.8, 36.1, 36.4, 36.7, 37.0},
            every axis y label/.style={font=\footnotesize,at={(current axis.north west)},right=2mm,above=0mm},
            legend style={fill=none,font=\small,at={(0.02,0.99)},anchor=north west,draw=none},
        ]
        \addplot[line width=0.3mm, mark=o,color=orange]
            plot coordinates {
                (1, 35.90)
                (2, 35.88)
                (3, 35.88)
                (4, 35.88)
                (5, 36.95)
            };
        \end{axis}

        \begin{axis}[
            height=\columnwidth/2.5,
            width=\columnwidth/2.1,
            ylabel={\it Micro-F1},
            xmin=0.5, xmax=5.5,
            ymin=50.0, ymax=52.0,
            axis y line*=right,
            ylabel near ticks,
            ytick={50.0, 50.5, 51.0, 51.5, 52.0},
            yticklabel style = {font=\scriptsize},
            xmajorticks=false,
            yticklabels={50.0, 50.5, 51.0, 51.5, 52.0},
            enlargelimits=false,
            every axis y label/.style={font=\footnotesize,at={(current axis.north west)},right=17mm,above=0mm},
            legend style={fill=none,font=\small,at={(0.02,0.99)},anchor=north west,draw=none},
        ]
        \addplot[line width=0.3mm, mark=square,color=blue]
            plot coordinates {
                (1, 50.24)
                (2, 50.28)
                (3, 50.28)
                (4, 50.28)
                (5, 51.68)
            };
        \end{axis}
    \end{tikzpicture}
    \hspace{2mm}
}
\subfloat[Amazon-Apps]{
    \begin{tikzpicture}[scale=1,every mark/.append style={mark size=2pt}]
        \begin{axis}[
            height=\columnwidth/2.5,
            width=\columnwidth/2.1,
            ylabel={\it Macro-F1},
            xmin=0.5, xmax=5.5,
            ymin=44.4, ymax=46,
            xtick={1,2,3,4,5},
            ytick={44.4, 44.8, 45.2, 45.6, 46},
            yticklabel style = {font=\scriptsize},
            xticklabel style = {font=\scriptsize},
            xticklabels={0.0, 0.5, 1.0, 1.5, 2.0},
            yticklabels={44.4, 44.8, 45.2, 45.6, 46},
            every axis y label/.style={font=\footnotesize,at={(current axis.north west)},right=2mm,above=0mm},
            legend style={fill=none,font=\small,at={(0.02,0.99)},anchor=north west,draw=none},
        ]
        \addplot[line width=0.3mm, mark=o,color=orange]
            plot coordinates {
                (1, 44.85)
                (2, 45.91)
                (3, 45.91)
                (4, 44.43)
                (5, 45.92)
            };
        \end{axis}

        \begin{axis}[
            height=\columnwidth/2.5,
            width=\columnwidth/2.1,
            ylabel={\it Micro-F1},
            xmin=0.5, xmax=5.5,
            ymin=57.5, ymax=61.5,
            axis y line*=right,
            ylabel near ticks,
            ytick={57.5, 58.5, 59.5, 60.5, 61.5},
            yticklabel style = {font=\scriptsize},
            xmajorticks=false,
            yticklabels={57.5, 58.5, 59.5, 60.5, 61.5},
            enlargelimits=false,
            every axis y label/.style={font=\footnotesize,at={(current axis.north west)},right=17mm,above=0mm},
            legend style={fill=none,font=\small,at={(0.02,0.99)},anchor=north west,draw=none},
        ]
        \addplot[line width=0.3mm, mark=square,color=blue]
            plot coordinates {
                (1, 57.79)
                (2, 61.05)
                (3, 61.05)
                (4, 58.77)
                (5, 61.06)
            };
        \end{axis}
    \end{tikzpicture}
    \hspace{2mm}
}
\subfloat[Goodreads-Poetry]{
    \begin{tikzpicture}[scale=1,every mark/.append style={mark size=2pt}]
        \begin{axis}[
            height=\columnwidth/2.5,
            width=\columnwidth/2.1,
            ylabel={\it Macro-F1},
            xmin=0.5, xmax=5.5,
            ymin=39.8, ymax=41.8,
            xtick={1,2,3,4,5},
            ytick={39.8, 40.2, 40.6, 41.0, 41.4, 41.8},
            yticklabel style = {font=\scriptsize},
            xticklabel style = {font=\scriptsize},
            xticklabels={0.0, 0.5, 1.0, 1.5, 2.0},
            yticklabels={39.8, 40.2, 40.6, 41.0, 41.4, 41.8},
            every axis y label/.style={font=\footnotesize,at={(current axis.north west)},right=2mm,above=0mm},
            legend style={fill=none,font=\small,at={(0.02,0.99)},anchor=north west,draw=none},
        ]
        \addplot[line width=0.3mm, mark=o,color=orange]
            plot coordinates {
                (1, 41.01)
                (2, 40.69)
                (3, 40.32)
                (4, 41.55)
                (5, 40.06)
            };
        \end{axis}

        \begin{axis}[
            height=\columnwidth/2.5,
            width=\columnwidth/2.1,
            ylabel={\it Micro-F1},
            xmin=0.5, xmax=5.5,
            ymin=50, ymax=52.5,
            axis y line*=right,
            ylabel near ticks,
            ytick={50, 50.5, 51, 51.5, 52, 52.5},
            yticklabel style = {font=\scriptsize},
            xmajorticks=false,
            yticklabels={50, 50.5, 51, 51.5, 52, 52.5},
            enlargelimits=false,
            every axis y label/.style={font=\footnotesize,at={(current axis.north west)},right=17mm,above=0mm},
            legend style={fill=none,font=\small,at={(0.02,0.99)},anchor=north west,draw=none},
        ]
        \addplot[line width=0.3mm, mark=square,color=blue]
            plot coordinates {
                (1, 52.31)
                (2, 51.98)
                (3, 50.81)
                (4, 52.44)
                (5, 50.49)
            };
        \end{axis}
    \end{tikzpicture}
    \hspace{2mm}
}
\subfloat[Google-Vermont]{
    \begin{tikzpicture}[scale=1,every mark/.append style={mark size=2pt}]
        \begin{axis}[
            height=\columnwidth/2.5,
            width=\columnwidth/2.1,
            ylabel={\it Macro-F1},
            xmin=0.5, xmax=5.5,
            ymin=56.2, ymax=58.0,
            xtick={1,2,3,4,5},
            ytick={56.2, 56.8, 57.4, 58.0},
            yticklabel style = {font=\scriptsize},
            xticklabel style = {font=\scriptsize},
            xticklabels={0.0, 0.5, 1.0, 1.5, 2.0},
            yticklabels={56.2, 56.8, 57.4, 58.0},
            every axis y label/.style={font=\footnotesize,at={(current axis.north west)},right=2mm,above=0mm},
            legend style={fill=none,font=\small,at={(0.02,0.99)},anchor=north west,draw=none},
        ]
        \addplot[line width=0.3mm, mark=o,color=orange]
            plot coordinates {
                (1, 57.82)
                (2, 57.02)
                (3, 56.64)
                (4, 57.77)
                (5, 56.68)
            };
        \end{axis}

        \begin{axis}[
            height=\columnwidth/2.5,
            width=\columnwidth/2.1,
            ylabel={\it Micro-F1},
            xmin=0.5, xmax=5.5,
            ymin=72.5, ymax=74,
            axis y line*=right,
            ylabel near ticks,
            ytick={72.5, 73, 73.5, 74},
            yticklabel style = {font=\scriptsize},
            xmajorticks=false,
            yticklabels={72.5, 73, 73.5, 74},
            enlargelimits=false,
            every axis y label/.style={font=\footnotesize,at={(current axis.north west)},right=17mm,above=0mm},
            legend style={fill=none,font=\small,at={(0.02,0.99)},anchor=north west,draw=none},
        ]
        \addplot[line width=0.3mm, mark=square,color=blue]
            plot coordinates {
                (1, 73.79)
                (2, 73.24)
                (3, 73.32)
                (4, 73.81)
                (5, 72.96)
            };
        \end{axis}
    \end{tikzpicture}
    \hspace{2mm}
}
\end{small}
\vspace{-2ex}
\caption{Macro-F1 and Micro-F1 when varying $\delta$} \label{fig:vary-delta}
\vspace{-2ex}
\end{figure*}
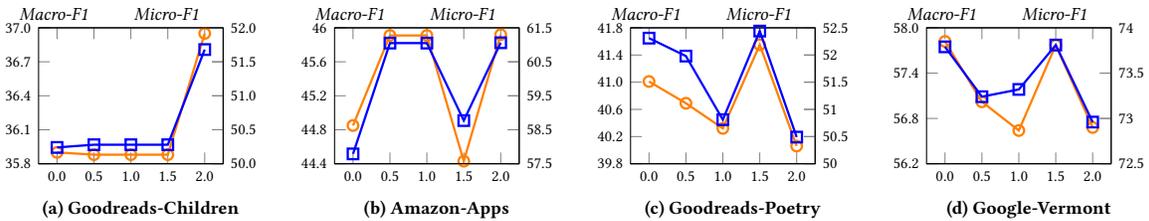
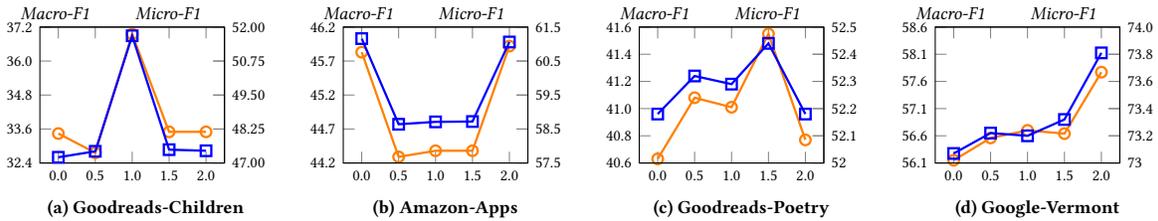
\begin{figure*}[!t]
\centering
\begin{small}
\subfloat[Goodreads-Children]{
    \begin{tikzpicture}[scale=1,every mark/.append style={mark size=2pt}]
        \begin{axis}[
            height=\columnwidth/2.5,
            width=\columnwidth/2.1,
            ylabel={\it Macro-F1},
            xmin=0.5, xmax=5.5,
            ymin=32.4, ymax=37.2,
            xtick={1,2,3,4,5},
            ytick={32.4, 33.6, 34.8, 36.0, 37.2},
            yticklabel style = {font=\scriptsize},
            xticklabel style = {font=\scriptsize},
            xticklabels={0.0, 0.5, 1.0, 1.5, 2.0},
            yticklabels={32.4, 33.6, 34.8, 36.0, 37.2},
            every axis y label/.style={font=\footnotesize,at={(current axis.north west)},right=2mm,above=0mm},
            legend style={fill=none,font=\small,at={(0.02,0.99)},anchor=north west,draw=none},
        ]
        \addplot[line width=0.3mm, mark=o,color=orange]
            plot coordinates {
                (1, 33.44)
                (2, 32.76)
                (3, 36.95)
                (4, 33.50)
                (5, 33.50)
            };
        \end{axis}

        \begin{axis}[
            height=\columnwidth/2.5,
            width=\columnwidth/2.1,
            ylabel={\it Micro-F1},
            xmin=0.5, xmax=5.5,
            ymin=47.00, ymax=52.00,
            axis y line*=right,
            ylabel near ticks,
            ytick={47.00, 48.25, 49.50, 50.75, 52.00},
            yticklabel style = {font=\scriptsize},
            xmajorticks=false,
            yticklabels={47.00, 48.25, 49.50, 50.75, 52.00},
            enlargelimits=false,
            every axis y label/.style={font=\footnotesize,at={(current axis.north west)},right=17mm,above=0mm},
            legend style={fill=none,font=\small,at={(0.02,0.99)},anchor=north west,draw=none},
        ]
        \addplot[line width=0.3mm, mark=square,color=blue]
            plot coordinates {
                (1, 47.21)
                (2, 47.43)
                (3, 51.68)
                (4, 47.49)
                (5, 47.45)
            };
        \end{axis}
    \end{tikzpicture}
    \hspace{2mm}
}
\subfloat[Amazon-Apps]{
    \begin{tikzpicture}[scale=1,every mark/.append style={mark size=2pt}]
        \begin{axis}[
            height=\columnwidth/2.5,
            width=\columnwidth/2.1,
            ylabel={\it Macro-F1},
            xmin=0.5, xmax=5.5,
            ymin=44.2, ymax=46.2,
            xtick={1,2,3,4,5},
            ytick={44.2, 44.7, 45.2, 45.7, 46.2},
            yticklabel style = {font=\scriptsize},
            xticklabel style = {font=\scriptsize},
            xticklabels={0.0, 0.5, 1.0, 1.5, 2.0},
            yticklabels={44.2, 44.7, 45.2, 45.7, 46.2},
            every axis y label/.style={font=\footnotesize,at={(current axis.north west)},right=2mm,above=0mm},
            legend style={fill=none,font=\small,at={(0.02,0.99)},anchor=north west,draw=none},
        ]
        \addplot[line width=0.3mm, mark=o,color=orange]
            plot coordinates {
                (1, 45.83)
                (2, 44.29)
                (3, 44.38)
                (4, 44.38)
                (5, 45.92)
            };
        \end{axis}

        \begin{axis}[
            height=\columnwidth/2.5,
            width=\columnwidth/2.1,
            ylabel={\it Micro-F1},
            xmin=0.5, xmax=5.5,
            ymin=57.5, ymax=61.5,
            axis y line*=right,
            ylabel near ticks,
            ytick={57.5, 58.5, 59.5, 60.5, 61.5},
            yticklabel style = {font=\scriptsize},
            xmajorticks=false,
            yticklabels={57.5, 58.5, 59.5, 60.5, 61.5},
            enlargelimits=false,
            every axis y label/.style={font=\footnotesize,at={(current axis.north west)},right=17mm,above=0mm},
            legend style={fill=none,font=\small,at={(0.02,0.99)},anchor=north west,draw=none},
        ]
        \addplot[line width=0.3mm, mark=square,color=blue]
            plot coordinates {
                (1, 61.16)
                (2, 58.64)
                (3, 58.71)
                (4, 58.72)
                (5, 61.06)
            };
        \end{axis}
    \end{tikzpicture}
    \hspace{2mm}
}
\subfloat[Goodreads-Poetry]{
    \begin{tikzpicture}[scale=1,every mark/.append style={mark size=2pt}]
        \begin{axis}[
            height=\columnwidth/2.5,
            width=\columnwidth/2.1,
            ylabel={\it Macro-F1},
            xmin=0.5, xmax=5.5,
            ymin=40.6, ymax=41.6,
            xtick={1,2,3,4,5},
            ytick={40.6, 40.8, 41.0, 41.2, 41.4, 41.6},
            yticklabel style = {font=\scriptsize},
            xticklabel style = {font=\scriptsize},
            xticklabels={0.0, 0.5, 1.0, 1.5, 2.0},
            yticklabels={40.6, 40.8, 41.0, 41.2, 41.4, 41.6},
            every axis y label/.style={font=\footnotesize,at={(current axis.north west)},right=2mm,above=0mm},
            legend style={fill=none,font=\small,at={(0.02,0.99)},anchor=north west,draw=none},
        ]
        \addplot[line width=0.3mm, mark=o,color=orange]
            plot coordinates {
                (1, 40.63)
                (2, 41.08)
                (3, 41.01)
                (4, 41.55)
                (5, 40.77)
            };
        \end{axis}

        \begin{axis}[
            height=\columnwidth/2.5,
            width=\columnwidth/2.1,
            ylabel={\it Micro-F1},
            xmin=0.5, xmax=5.5,
            ymin=52, ymax=52.5,
            axis y line*=right,
            ylabel near ticks,
            ytick={52, 52.1, 52.2, 52.3, 52.4, 52.5},
            yticklabel style = {font=\scriptsize},
            xmajorticks=false,
            yticklabels={52, 52.1, 52.2, 52.3, 52.4, 52.5},
            enlargelimits=false,
            every axis y label/.style={font=\footnotesize,at={(current axis.north west)},right=17mm,above=0mm},
            legend style={fill=none,font=\small,at={(0.02,0.99)},anchor=north west,draw=none},
        ]
        \addplot[line width=0.3mm, mark=square,color=blue]
            plot coordinates {
                (1, 52.18)
                (2, 52.32)
                (3, 52.29)
                (4, 52.44)
                (5, 52.18)
            };
        \end{axis}
    \end{tikzpicture}
    \hspace{2mm}
}
\subfloat[Google-Vermont]{
    \begin{tikzpicture}[scale=1,every mark/.append style={mark size=2pt}]
        \begin{axis}[
            height=\columnwidth/2.5,
            width=\columnwidth/2.1,
            ylabel={\it Macro-F1},
            xmin=0.5, xmax=5.5,
            ymin=56.1, ymax=58.6,
            xtick={1,2,3,4,5},
            ytick={56.1, 56.6, 57.1, 57.6, 58.1, 58.6},
            yticklabel style = {font=\scriptsize},
            xticklabel style = {font=\scriptsize},
            xticklabels={0.0, 0.5, 1.0, 1.5, 2.0},
            yticklabels={56.1, 56.6, 57.1, 57.6, 58.1, 58.6},
            every axis y label/.style={font=\footnotesize,at={(current axis.north west)},right=2mm,above=0mm},
            legend style={fill=none,font=\small,at={(0.02,0.99)},anchor=north west,draw=none},
        ]
        \addplot[line width=0.3mm, mark=o,color=orange]
            plot coordinates {
                (1, 56.15)
                (2, 56.56)
                (3, 56.70)
                (4, 56.64)
                (5, 57.77)
            };
        \end{axis}

        \begin{axis}[
            height=\columnwidth/2.5,
            width=\columnwidth/2.1,
            ylabel={\it Micro-F1},
            xmin=0.5, xmax=5.5,
            ymin=73, ymax=74.0,
            axis y line*=right,
            ylabel near ticks,
            ytick={73, 73.2, 73.4, 73.6, 73.8, 74.0},
            yticklabel style = {font=\scriptsize},
            xmajorticks=false,
            yticklabels={73, 73.2, 73.4, 73.6, 73.8, 74.0},
            enlargelimits=false,
            every axis y label/.style={font=\footnotesize,at={(current axis.north west)},right=17mm,above=0mm},
            legend style={fill=none,font=\small,at={(0.02,0.99)},anchor=north west,draw=none},
        ]
        \addplot[line width=0.3mm, mark=square,color=blue]
            plot coordinates {
                (1, 73.07)
                (2, 73.22)
                (3, 73.20)
                (4, 73.32)
                (5, 73.81)
            };
        \end{axis}
    \end{tikzpicture}
    \hspace{2mm}
}
\end{small}
\vspace{-2ex}
\caption{Macro-F1 and Micro-F1 when varying $\lambda$} \label{fig:vary-lambda}
\vspace{-2ex}
\end{figure*}

\begin{table*}[!t]
\centering
\caption{Hyper-parameters of \algognn.}
\resizebox{0.9\textwidth}{!}{%
\fontsize{15}{8}\selectfont
\begin{tabular}{l c c c c c c c c c c c}
\toprule
\textbf{dataset} & \textbf{learning rate} & \textbf{weight decay} & $\boldsymbol{\epsilon}$ & \textbf{epochs} & \textbf{early stop} & \textbf{batch size} & \textbf{MP layer} & \textbf{SVD dim}   & $\boldsymbol{\delta}$ & $\boldsymbol{\lambda}$ & \textbf{PLM} \\
\midrule
Goodreads-Children & 1e-3 & 1e-2 & 1e-6 & 300 & 30 & full batch & 3& 256  & 2& 1 & BERT-Tiny \\
Amazon-Apps & 1e-3 & 1e-2 & 1e-6 & 300 & 30 & full batch & 1& 64  & 2& 2& BERT-Tiny \\
Amazon-Movie & 1e-3 & 1e-2 & 1e-6 & 300 & 30 & full batch & 1& 16  & 0& 0& BERT-Tiny \\
Goodreads-Crime & 1e-3 & 1e-2 & 1e-6 & 300 & 30 & full batch & 2 & 64 & 1& 1& BERT-Tiny \\
\midrule
Goodreads-Poetry & 1e-5 & 1e-3 & 1e-8 & 100 & 3 & 25 & 4 & 64 & 1.5& 1.5&BERT-Base \\
Google-Vermont & 1e-5 & 1e-3 & 1e-8 & 100 & 3 & 25 & 2 & 64 & 1.5& 2&BERT-Base \\
Google-Hawaii & 1e-5 & 1e-3 & 1e-8 & 100 & 3 & 25 & 2 & 64 & 1& 0&BERT-Base \\
Amazon-Products & 1e-5 & 1e-3 & 1e-8 & 100 & 3 & 25 & 2 & 64 & 1.5& 2&BERT-Base \\
\bottomrule
\end{tabular}%
}
\label{tab:hyperparameters_gnn}
\end{table*}

\begin{table*}[!t]
\centering
\caption{Hyper-parameters of \algogau.}
\resizebox{0.9\textwidth}{!}{%
\fontsize{15}{8}\selectfont
\begin{tabular}{l c c c c c c c c c c c}
\toprule
\textbf{dataset} & \textbf{learning rate} & \textbf{weight decay} & $\boldsymbol{\epsilon}$ & \textbf{epochs} & \textbf{early stop} & \textbf{batch size} & \textbf{MP layer} & \textbf{SVD dim}   & $\boldsymbol{\delta}$ & $\boldsymbol{\lambda}$ & \textbf{PLM} \\
\midrule
Goodreads-Children & 1e-3 & 1e-2 & 1e-6 & 300 & 30 & full batch & 2& 64  & 0& 0.5& BERT-Tiny \\
Amazon-Apps & 1e-3 & 1e-2& 1e-6 & 300 & 30 & full batch & 2& 64  & 0& 0& BERT-Tiny \\
Amazon-Movie & 1e-3 & 1e-2& 1e-6 & 300 & 30 & full batch & 5& 32  & 1.5& 1.5& BERT-Tiny \\
Goodreads-Crime & 1e-3 & 1e-2& 1e-6 & 300 & 30 & full batch & 4 & 64 & 2& 1.5& BERT-Tiny \\
\midrule
Goodreads-Poetry & 1e-5 & 1e-3& 1e-8 & 100 & 3 & 25 & 2 & 64 & 1.5& 0.5&BERT-Base \\
Google-Vermont & 1e-5 & 1e-3& 1e-8 & 100 & 3 & 25 & 3 & 64 & 2& 1.5&BERT-Base \\
Google-Hawaii & 1e-5 & 1e-3& 1e-8 & 100 & 3 & 25 & 5 & 64 & 0.5& 0.5&BERT-Base \\
Amazon-Products & 1e-5 & 1e-3& 1e-8 & 100 & 3 & 25 & 1 & 64 & 0& 0&BERT-Base \\
\bottomrule
\end{tabular}%
}
\label{tab:hyperparameters_gau}
\end{table*}

\subsection{Datasets}
The TINs utilized in this work are drawn from three real-world TINs: Amazon~\cite{he2016ups}, Goodreads~\cite{wan2019fine}, and Google~\cite{li2022uctopic}. Amazon is a user-item TIN, where reviews are treated as text on interactions; Goodreads is a reader-book TIN, where readers’ comments are used as interaction text information; Google is a business-review TIN, which contains review information on Google map, business metadata, and links, and these reviews are regarded as attributes between interaction connection. Since Amazon and Goodreads both have multiple domains and Google has many states in the United States, we select three domains for each of Amazon and Goodreads and two states for Google. There are 5 categories for interactions in Amazon (i.e., 1-star, ..., 5-star), 6 categories for interactions in Goodreads (i.e., 0-star, ..., 5-star) and 5 categories for interactions in Google (i.e., 1-star, ..., 5-star), where the interaction labels represent users’ ratings on the business entities. 
The eight most important selected TINs are Goodreads-Children, Amazon-Apps, Amazon-Movie, Goodreads-Crime, Goodreads-Poetry, Google-Vermont, Google-Hawaii, and Amazon-Products. Refer to Table \ref{tab:dataset_statistics} for the dataset statistics. For small TINs Goodreads-Children, Amazon-Apps, Amazon-Movie, and Goodreads-Crime, we choose BERT-Tiny to do the full-batch training. For the rest TINs, we pick BERT-Base to do the mini-batch training. 

\subsection{Baselines}
The baselines considered in our study can be categorized into three groups: bag-of-words methods, pretrained language models, and edge-wise representation learning methods. In Table \ref{tab:baseline_codes}, we list the available URL of each method. 

\stitle{Bag-of-Words Method} We use TF-IDF~\cite{robertson1994some} as a representative bag-of-words method. This method captures the term frequency-inverse document frequency to quantify the importance of words in the context of the document. To further enhance the performance, we concatenate node embeddings with the TF-IDF vector (TF-IDF+nodes), thereby incorporating TIN information.

\stitle{Pretrained Language Model} We use BERT~\cite{devlin2018bert} as the baseline pretrained language model. BERT enables contextual representation of the text, which is crucial for capturing semantic relationships. We further extend BERT by incorporating TIN information, appending node embeddings to the input token sequence (BERT+nodes). Graphformers uses PLM tnlrv3, an alias of UniLM v2, which is available internally in Microsoft at the current stage, so we use BERT instead of tnlrv3 to do the Graphformers baseline. 

\stitle{Edge-Wise Representation Learning Methods} The third category of baselines consists of edge-wise representation learning methods, which include AttrE2Vec~\cite{bielak2022attre2vec}, TER+AER~\cite{wang2023efficient}, and EAGLE~\cite{wang2024effective}. These methods specifically learn edge representations by considering both structural and attribute information. For TER+AER and EAGLE, we adopt the hyperparameters mentioned in their respective papers and use pretrained BERT embeddings as the initial features for edges.

\subsection{Hyperparameters}
Table \ref{tab:hyperparameters_gnn} and Table \ref{tab:hyperparameters_gau} present the key hyperparameters used in the model. \textbf{Learning rate}, \textbf{weight decay}, $\boldsymbol{\beta_1}$ (first-moment decay rate), $\boldsymbol{\beta_2}$ (second-moment decay rate), and $\boldsymbol{\epsilon}$ (numerical stability constant) are key parameters of the AdamW optimizer, controlling the learning rate, adaptive momentum updates, and regularization. By default, $\boldsymbol{\beta_1}$ is set to 0.9, and $\boldsymbol{\beta_2}$ is set to 0.999, and thus we do not show them in the table. \textbf{Early stopping} specifies the number of epochs without improvement before halting training. \textbf{Batch size} indicates the number of samples processed before updating the model weights and \textbf{full batch} means taking the whole TIN as the input. \textbf{MP layer} denotes the number of layers in the LGA or GAU, controlling the model depth and learning capacity. \textbf{SVD dim} represents the dimensionality of the truncated SVD of our Structural Encoding for Interactions. \textbf{PLM} refers to the specific pretrained language model used, such as \textbf{BERT-Tiny} or \textbf{BERT-Base}.

\subsection{Hyperparameter Analysis}
\subsubsection{\bf Effect of Varying Number of LGA/GAU Layers}
As shown in Figure~\ref{fig:vary-layer}, varying the number of layers from 1 to 5 reveals distinct patterns across different TINs. For Goodreads-Children, performance, measured by both Macro-F1 and Micro-F1, improves up to 3 layers, after which a decline is observed, suggesting overfitting in deeper architectures. Similarly, Amazon-Apps achieves its best performance at 2 layers, indicating that a shallow architecture suffices to capture meaningful interactions. In contrast, Goodreads-Poetry reaches its optimal performance with 4 layers, while Google-Vermont exhibits peak performance at 2 layers, suggesting that each TIN benefits from varying LGA/GAU depths to capture relevant structural patterns.
\subsubsection{\bf Effect of Varying SVD Embedding Dimensionality}
Figure~\ref{fig:vary-dim} presents the influence of embedding dimensionality, derived from the truncated SVD in our Structural Encoding for Interactions, on TIC performance. The results indicate that Goodreads-Children benefits most from 256 dimensions, while Amazon-Apps achieves its highest performance at 64 dimensions. For Goodreads-Poetry, the optimal dimensionality is found to be 128, whereas Google-Vermont shows a preference for 64 dimensions. These results suggest that higher dimensions may capture more complex relationships in certain TINs, but may lead to overfitting if increased beyond optimal levels.
\subsubsection{\bf Effect of Varying Parameter $\boldsymbol{\delta}$}
The effect of varying the parameter $\boldsymbol{\delta}$ is illustrated in Figure~\ref{fig:vary-delta}. For Goodreads-Children and Amazon-Apps, the best performance is observed at $\boldsymbol{\delta} = 2.0$. In contrast, both Goodreads-Poetry and Google-Vermont perform optimally at $\boldsymbol{\delta} = 1.5$, suggesting that these TINs may benefit from more moderate values, potentially avoiding overfitting by controlling the impact of the parameter on model complexity.
\subsubsection{\bf Effect of Varying Parameter $\boldsymbol{\lambda}$}
Figure~\ref{fig:vary-lambda} presents the effect of varying $\boldsymbol{\lambda}$ on TIC performance. For Goodreads-Children, the optimal value of $\boldsymbol{\lambda}$ is 1.0, indicating that balancing the residual connection at this level is most effective. Amazon-Apps, on the other hand, reaches its best performance at $\boldsymbol{\lambda} = 2.0$, suggesting a stronger residual effect is beneficial. Similarly, Goodreads-Poetry performs best at $\boldsymbol{\lambda} = 1.5$, while Google-Vermont continues to improve with increasing values of $\boldsymbol{\lambda}$, achieving its highest results at $\boldsymbol{\lambda} = 2.0$. These variations across TINs imply that the residual connection parameter needs to be carefully tuned to control the impact on feature integration while preserving the model's ability to generalize.

\section{Theoretical Proofs}\label{sec:proof}

\begin{proof}[\bf Proof of Lemma~\ref{lem:softmax-P}]
First, according to the definitions of $\PUM$ and $\PVM$ in Eq.~\eqref{eq:PUV}, we can get
\begin{small}
\begin{equation*}
\begin{split}
\PUM_{e_{u,i},e_{u,l}} = (\EUM\DUM^{-1}\EUM^{\top})_{e_{u,i},e_{u,l}}=\frac{1}{\dvec_{u}+1},\\ 
\PVM_{e_{u,i},e_{v,i}} = (\EVM\DVM^{-1}\EVM^{\top})_{e_{u,i},e_{v,i}}=\frac{1}{\dvec_{i}+1}.
\end{split}
\end{equation*}
\end{small}
Next, by the sparse softmax operations, we can derive
\begin{small}
\begin{align*}
\textsf{ssoftmax}\left(\EUM\EUM^{\top}/\sqrt{|\U|}\right)_{e_{u,i},e_{i,l}} & = \frac{exp(1/\sqrt{|\U|})}{(\dvec_{u}+1)\cdot exp(1/\sqrt{|\U|})}\\
& =\frac{1}{\dvec_{u}+1} = \PUM_{e_{u,i},e_{u,l}}\\
\textsf{ssoftmax}\left(\EVM\EVM^{\top}/\sqrt{|\V|}\right)_{e_{u,i},e_{l,j}} & = \frac{exp(1/\sqrt{|\V|})}{(\dvec_{i}+1)\cdot exp(1/\sqrt{|\V|})}\\
& =\frac{1}{\dvec_{i}+1} = \PVM_{e_{u,i},e_{v,i}},
\end{align*}
\end{small}
which completes the proof.
\end{proof}

\begin{proof}[\bf Proof of Theorem~\ref{lem:Lap-Inv}]
\begin{lemma}\label{lem:Lap-P}
The Laplacian matrix $\widetilde{\LM}$ of $\widetilde{\G}$ is $\IM-\PM$.
\end{lemma}

By Lemma~\ref{lem:Lap-P}, $\widetilde{\LM}=\IM-\PM$. Since $\DM^{-\frac{1}{2}}{\EM}=\UM\boldsymbol{\Sigma}\VM^{\top}$ and singular vectors $\UM$ satisfy $\UM^{\top}\UM=\IM$, 
\begin{align*}
\widetilde{\LM} = \IM-\PM & = \IM - \frac{1}{2}\VM\boldsymbol{\Sigma}\UM^{\top} \cdot \UM\boldsymbol{\Sigma}\VM^{\top} \\
& = \IM - \VM{\frac{\boldsymbol{\Sigma}^2}{2}}\VM^{\top} = \VM (\IM - \boldsymbol{\Sigma}^2/2) \VM^{\top}.
\end{align*}
The pseudo-inverse $\widetilde{\LM}^{\dagger}$ of $\widetilde{\LM}$, i.e., $(\IM-\PM)^{-1}$, can be represented by
\begin{equation*}
(\VM (\IM - \boldsymbol{\Sigma}^2/2) \VM^{\top})^{-1} = \VM (\IM - \boldsymbol{\Sigma}^2/2)^{-1} \VM^{\top} = \VM \frac{1}{\IM - \boldsymbol{\Sigma}^2/2} \VM^{\top}.
\end{equation*}

By the definition of $\ZM^{(\textrm{d})}$ in Eq.~\eqref{eq:dist-emb}, it is easy to verify that $\ZM^{(\textrm{d})}{\ZM^{(\textrm{d})}}^{\top}=\widetilde{\LM}^{\dagger}$. As a consequence, $\forall{e_{u,i}, e_{u,j}\in \EDG}$, 
\begin{align*}
& \|\ZM^{(\textrm{d})}_{e_{u,i}} -\ZM^{(\textrm{d})}_{e_{u,j}}\|^2 = \|{\ZM^{(\textrm{d})}}^{\top}\cdot (\mathbf{1}_{e_{u,i}}-\mathbf{1}_{e_{u,i}})\|^2\\
& = (\mathbf{1}_{e_{u,i}}-\mathbf{1}_{e_{u,i}})^{\top}\ZM^{(\textrm{d})}{\ZM^{(\textrm{d})}}^{\top}(\mathbf{1}_{e_{u,i}}-\mathbf{1}_{e_{u,i}}) \\
& = (\mathbf{1}_{e_{u,i}}-\mathbf{1}_{e_{u,j}})^{\top} \widetilde{\LM}^{\dagger} (\mathbf{1}_{e_{u,i}}-\mathbf{1}_{e_{u,j}}) = RD(e_{u,i},e_{u,j}),
\end{align*}
where $\mathbf{1}_{e_{u,i}}$ (resp. $\mathbf{1}_{e_{u,j}}$) be the unit vector with 1 at entry $e_{u,i}$ (resp. $e_{u,j}$) and $0$ everywhere else. The lemma is then proved.
\end{proof}

\begin{proof}[\bf Proof of Lemma~\ref{lem:Lap-P}]
We first prove that $\PM$ is a row-stochastic matrix. 
Since it is symmetric, then its doubly stochastic property naturally follows. By $\PM= \EM^{\top}\frac{\DM^{-1}}{2}\EM$ , the $(i,j)$-th entry $\PM_{e_{u,i}, e_{u,j}}$ of $\PM$ can be represented as: 
\begin{align*}
    \sum_{e_{u,j} \in \EDG} \PM_{e_{u,i}, e_{u,j}} = \sum_{e_{u,j} \in \EDG} \frac{1}{\DM_{u, u} \cdot \mathbb{1}_{u \in e_{u,j}}} = 1,
\end{align*}
where $\mathbb{1}_{u \in e_j}$ is an indicator function which equals 1 when node $u$ is an endpoint of edge $e_{u,j}$.
Consequently, the diagonal degree matrix of $\widetilde{\G}$ is then $\IM$.
According to the definition of graph Laplacian, the lemma is proved.
\end{proof}
\begin{proof}[\bf Proof of Theorem~\ref{lem:Katz}]
When $k=|\U|+|\V|$, $\UM\boldsymbol{\Sigma}{\VM}^{\top}$ is the full SVD of $\DM^{-\frac{1}{2}}{\EM}$, and hence, $\PM=\VM\frac{\boldsymbol{\Sigma}^{2}}{2}\VM^{\top}$. In turn, if $\alpha=1$,
\begin{align*}
\sum_{k=0}^{\infty}{\alpha^k\PM^k} & = \sum_{k=0}^{\infty}{\alpha^k\VM\frac{\boldsymbol{\Sigma}^{2k}}{2^k}\VM^{\top} }=  \VM \left(\sum_{k=0}^{\infty}{\alpha^k\frac{\boldsymbol{\Sigma}^{2k}}{2^k}}\right) \VM^{\top} \\
& = \VM \frac{1}{1-\alpha\boldsymbol{\Sigma}^2/2} \VM^{\top} = \VM \frac{1}{1-\boldsymbol{\Sigma}^2/2} \VM^{\top}= \ZM^{(\textrm{d})}{\ZM^{(\textrm{d})}}^{\top},
\end{align*}
which finishes the proof.
\end{proof}

\begin{proof}[\bf Proof of Theorem~\ref{lem:centrality}]
Let $\LM=\DM-\AM$ be the Laplacian of $\G$. According to \cite{peng2021local}, the spanning centrality $s(e_{u,i})$ is equivalent to 
\begin{equation*}
s(e_{u,i})=\LM^{\dagger}_{u,u}+\LM^{\dagger}_{i,i}-\LM^{\dagger}_{u,i}-\LM^{\dagger}_{i,u},
\end{equation*}
where $\LM^{\dagger}$ is the Moore-Penrose pseudo-inverse of $\LM$.
Since $\LM = \BM\BM^{\top}$ and $\PhiM\LaM\PsiM^{\top}$ is the SVD of $\BM$, it is easy to get that $$\LM = \PhiM\LaM\PsiM^{\top} \PsiM\LaM\PhiM^{\top} = \PhiM\LaM^2\PhiM^{\top}$$ is the eigendecomposition of $\LM$. By Section 2.2 in \cite{spielman2008graph}, $\LM^{\dagger}=\PhiM \LaM^{-2}\PhiM^{\top}$.
Therefore, $$\BM^{\top}\LM^{\dagger}\BM=\PsiM\LaM\PhiM^{\top} \cdot \PhiM \LaM^{-2}\PhiM^{\top} \cdot \PhiM\LaM\PsiM^{\top}=\PsiM\PsiM^{\top}.$$

First, $(\BM^{\top}\LM^{\dagger})_{e_{u,i},x}=\BM^{\top}_{e_{u,i}}\cdot \LM^{\dagger}_{\cdot,x}=\LM^{\dagger}_{u,x}-\LM^{\dagger}_{i,x}$, meaning that
\begin{equation*}
\begin{split}
(\BM^{\top}\LM^{\dagger})_{e_{u,i},u}=\LM^{\dagger}_{u,u}-\LM^{\dagger}_{i,u},\\(\BM^{\top}\LM^{\dagger})_{e_{u,i},i}=\LM^{\dagger}_{u,i}-\LM^{\dagger}_{i,i}.
\end{split}
\end{equation*}
Hence, $(\BM^{\top}\LM^{\dagger})_{e_{u,i},u}-(\BM^{\top}\LM^{\dagger})_{e_{u,i},i}=s(e_{u,i})$.
Then we can derive
\begin{align*}
(\BM^{\top}\LM^{\dagger}\BM)_{e_{u,i},e_{u,i}} & =(\BM^{\top}\LM^{\dagger})_{e_{u,i},u}-(\BM^{\top}\LM^{\dagger})_{e_{u,i},i} = s(e_{u,i})
\end{align*}
Recall that $\BM^{\top}\LM^{\dagger}\BM = \PsiM\PsiM^{\top}$.
This means $s(e_{u,i})=\PsiM_{e_{u,i}}\cdot \PsiM_{e_{u,i}}^{\top}=\|\PsiM_{e_{u,i}}\|^2_2$.
The theorem is proved.
\end{proof}

\balance

\end{document}